\documentclass[review]{elsarticle}
\usepackage{longfigure, multirow}
\usepackage[title,toc,page]{appendix}
\usepackage{subfloat, subcaption}
\usepackage{algorithm} 
\usepackage{algpseudocode} 
\usepackage{float}
\usepackage[table,xcdraw]{xcolor}
\usepackage[justification=centering]{caption}
\usepackage{amsmath,textcomp,amssymb,geometry,graphicx,enumerate}
\usepackage{natbib}
\usepackage{lineno, nccmath}       
\usepackage[hidelinks]{hyperref}
\usepackage{url, array}            
\usepackage{booktabs}       
\usepackage{amsfonts}       
\usepackage{dsfont}
\usepackage{microtype}      
\graphicspath{ {figures/} }

\journal{Journal of Accident Analysis \& Prevention}

\biboptions{authoryear}

\begin{document}

\begin{frontmatter}

\title{Connecting Surrogate Safety Measures to Crash Probablity via Causal Probabilistic Time Series Prediction}


\author[address1]{Jiajian Lu\corref{mycorrespondingauthor}}
\cortext[mycorrespondingauthor]{Corresponding author}
\ead{jiajian\_lu@berkeley.edu}

\author[address1]{Offer Grembek}
\ead{grembek@berkeley.edu}

\author[address2]{Mark Hansen}
\ead{mhansen@ce.berkeley.edu}

\address[address1]{Safe Transportation Research and Education Center, University of California, Berkeley, 2614 Dwight Way \#7374, Berkeley, CA 94720, U.S.A.}

\address[address2]{114 McLaughlin Hall, Institute of Transportation Studies, University of California, Berkeley, Berkeley, CA 94720, U.S.A.}

\begin{abstract}
Surrogate safety measures can provide fast and pro-active safety analysis and give insights on the pre-crash process and crash failure mechanism by studying near misses. However, validating surrogate safety measures by connecting them to crashes is still an open question. This paper proposed a method to connect surrogate safety measures to crash probability using probabilistic time series prediction. The method used sequences of speed, acceleration and time-to-collision to estimate the probability density functions of those variables with transformer masked autoregressive flow (transformer-MAF). The autoregressive structure mimicked the causal relationship between condition, action and crash outcome and the probability density functions are used to calculate the conditional action probability, crash probability and conditional crash probability. The predicted sequence is accurate and the estimated probability is reasonable under both traffic conflict context and normal interaction context and the conditional crash probability shows the effectiveness of evasive action to avoid crashes in a counterfactual experiment.
\end{abstract}

\begin{keyword}
Surrogate Safety Measure, Time to collision, Evasive Action, Crash Probability, Density Estimation, Transformer, Deep Unsupervised Learning, Counterfactual Experiment
\end{keyword}
\end{frontmatter}


\section{Introduction}
Traditional crash-based safety analysis has many limitations since crash data has small sample size which leads to unobserved heterogeneity \citep{lord2010statistical, mannering2014analytic}, lacks detailed information describing the crash process, and can improve traffic safety only after crashes happen \citep{laureshyn2016review, johnsson2018search}. On the other hand, traffic safety analysis using surrogate safety measures (SSM) have attracted more and more interest, since it can provide fast, pro-active safety analysis, while also yielding insights on the pre-crash process and crash failure mechanism by studying near misses \citep{tarko2009surrogate, tarko2019measuring}.

\begin{figure}[H]
	\centering
	\includegraphics[width=0.6\linewidth]{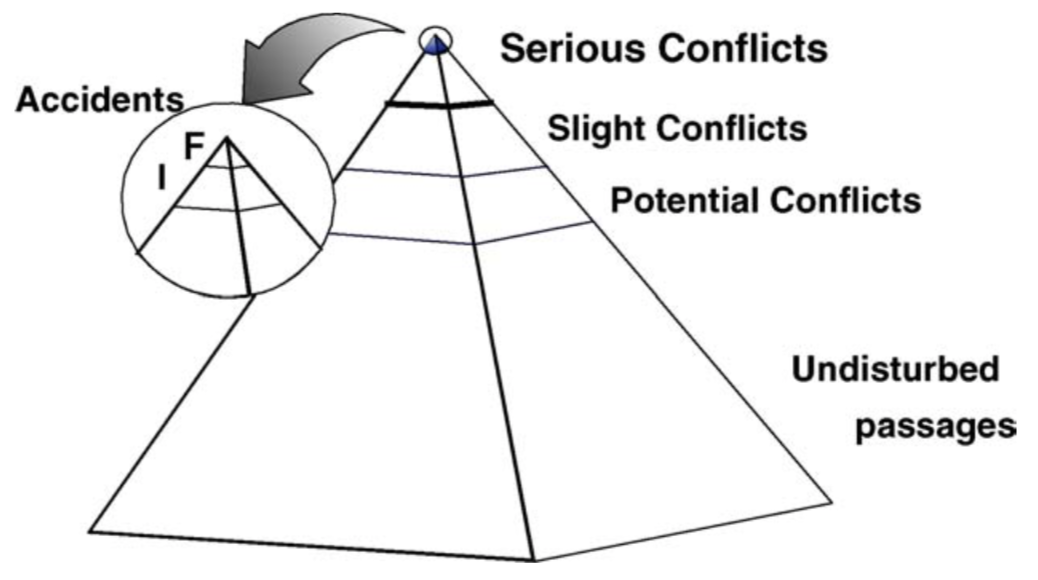}
	\caption{Safety Pyramid - the Interaction between Road Users as a Continuum of Events \citep{hyden1987development}}
	\label{fig:pyramid}
\end{figure}

The theoretical foundation of surrogate safety indicators assumes that all traffic events are related to safety. These traffic events have different degree of severity (unsafety) and a relationship exists between the severity and the frequency of events shown as Figure~\ref{fig:pyramid} \citep{hyden1987development}. The severity of an event is often measured by the proximity in space or time two road users. Time to collision (TTC) \citep{hayward1971near} and post-encroachment time (PET) \citep{allen1978analysis} are two most often used indicators. TTC is the time remaining before the collision if the involved road users continue with their respective speeds and trajectories and can be calculated as long when vehicles are on a collision course. The minimum TTC during an interaction is compared to a pre-defined threshold ($1.5s$ \citep{sacchi2013comparison}) to determine whether this event is a traffic conflict or a normal interaction.
If the relationships between the layers of the safety pyramid is known, it is theoretically possible to calculate the frequency of the very severe but infrequent events (accidents) based on the known frequency of the less severe but more frequently occurring events \citep{svensson2006estimating}. However, connecting traffic conflicts to crashes is still an open question and several methods have been proposed \citep{zheng2021modeling}. \citeauthor{hauer1986research} proposed a regression model to relate conflicts and crashes. \citeauthor{davis2011outline} used a structural equation to model causal relationships among initial condition, action and crash outcome to estimate the crash probability. \citeauthor{songchitruksa2006extreme} used the extreme value theory (EVT) to model the distribution of TTC and calculated the probability of TTC reaching the extreme level ($\text{TTC}=0$) as the crash probability. A more detailed review of these methods is in the next section.

With the development of deep unsupervised learning in computer science field, generative models \citep{kingma2014semi} can learn the distribution of the data and generate new samples that are similar to the original data. Some of the models have been applied in the transportation field. \citeauthor{chen2021traffic} used generative adversarial network (GAN) to generate traffic accident and \citeauthor{ding2020learning} used probability graphic model to generate safety-critical scenarios. \citeauthor{lu2022learning} used transformer encoder to learn the representation of time series of SSM data to identify traffic conflict. By incorporating neural networks, such as Long short-term memory (LSTM) and transformer, that can deal with time series data, probabilistic time series prediction models \citep{salinas2020deepar} can predict the distribution of the data every time step. Therefore, the distribution of time series of TTC can be estimated and the crash probability can be calculated with probabilistic time series prediction.

In this paper, we propose a non-crash-based method to relate surrogate safety measures to crashes based on the causal model. Our main contributions are:
\begin{enumerate}[1.]
\item The method uses transformer-MAF to predict real time crash probability by estimating the probability density function of surrogate safety measures for every time step.
\item The method implements the dependency structure among condition, action and crash outcome from the causal model into the probability density functions with an autoregressive network.
\item The method overcomes the limitations of the causal model and uses all values of condition, action and crash outcome by treating them as continuous variables .
\item We estimates the model on real-world traffic data to compare the crash probability under traffic conflict and normal interaction scenarios and calculate the effectiveness of evasive action.
\end{enumerate}
\section{Literature Review}
\subsection{Connecting SSMs to Crashes}
The regression-based method can directly model the relationship between traffic conflict and crashes given the count of both data. \citeauthor{hauer1982traffic} used linear regression model with form as $\lambda = \pi \cdot c$ where $\lambda$ is the number of crashes on an entity during a certain period of time, $c$ is the number of traffic conflicts on the same entity of the same time and $\pi$ is the crash-to-conflict ratio. \citep{el2013safety} estimated the counts of traffic conflict from traffic volume with Poisson-lognormal model and incorporated it in a safety performance function with negative binomial model. The regression-based model are easy to understand and apply but this approach still requires crash data which suffers from the same issues of the traditional road safety analysis and the crash-to-conflict ratios may vary for different road entities and time periods \citep{zheng2014traffic}.

\begin{figure}[H]
    \centering
    \begin{subfigure}{0.5\textwidth}
        \centering
        \includegraphics[width=\linewidth]{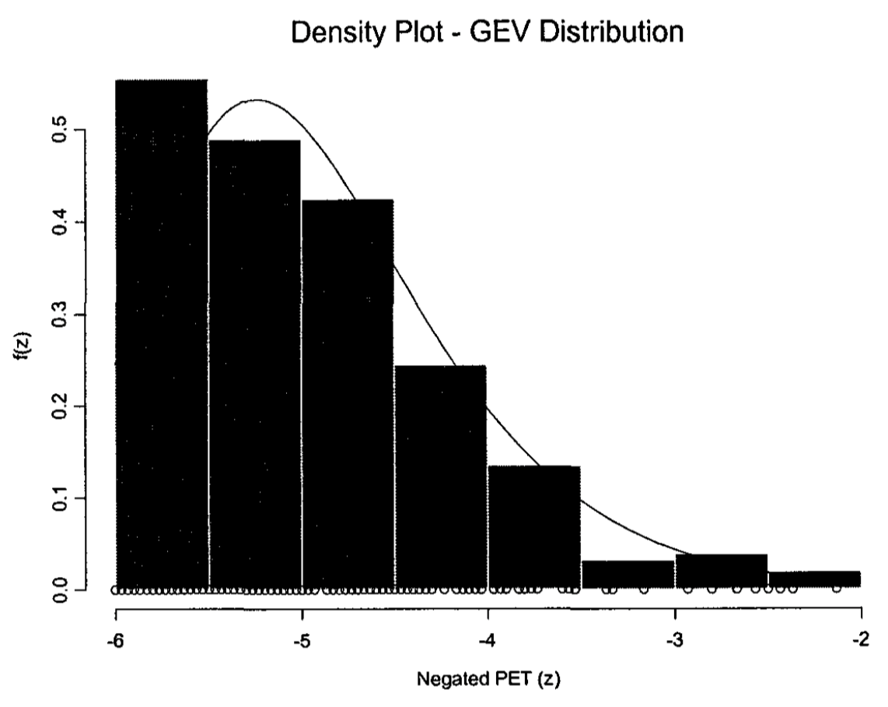}
        \caption{Distribution of TTC estimated from EVT model \citep{songchitruksa2004innovative}}
        \label{fig:evt}
    \end{subfigure}%
    \begin{subfigure}{0.5\textwidth}
        \centering
        \includegraphics[width=0.76\linewidth]{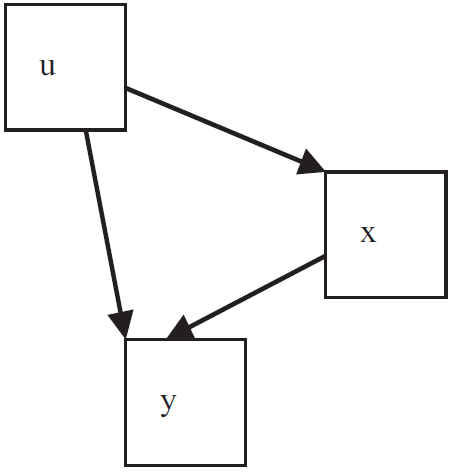}
        \caption{Dependency structure among initial condition $u$, action $x$ and crash outcome $y$ \citep{davis2011outline}}
        \label{fig:causal}
    \end{subfigure}%
\caption{Illustration of the EVT model and the causal model}\label{fig:lr_evt_cm}
\end{figure}

The EVT method can extrapolate the distribution of the observed traffic conflicts to the unobserved crashes to calculate the crash probability as shown in Figure~\ref{fig:evt}. Traffic conflicts are measured by SSMs like TTC and if TTC reaches the extreme level ($\text{TTC}=0$), traffic conflicts would become crashes. The risk of crash can be calculated as Equation~\ref{eq:gev} \useshortskip

\begin{align} \label{eq:gev}
    R=\operatorname{Pr}(Z \geq 0)=1-G(0)
\end{align}
where $R$ is the risk of crash, $Z$ is the negated TTC, and $G(\cdot)$ is the generalized extreme value distribution or the generalized Pareto distribution. There is growing interest in using EVT for traffic conflict-based safety estimation through the application of advanced statistical methods. \citep{zheng2018bivariate, zheng2019univariate} used bivariate generalized Pareto distribution to estimate crashes with several different SSMs. With the combined use of different indicators, the model provides a more holistic approach to measure the severity of an event. \citep{zheng2019application, zheng2019bayesian} developed Bayesian hierarchical extreme value models to combine traffic conflicts from different sites for crash estimation in order to overcome the problem that severe traffic conflicts are rare for each individual site. However, the traffic conflict indicators are mainly proximity metrics such as TTC and PET while evasive action-based indicators are overlooked. Moreover, the statistical models have their inherent model assumptions like the parameters of GEV distribution are linearly related to the site properties \citep{fu2021random} that sometimes the data do not follow. Additionally, indicators used in EVT are the extreme of a sequence of SSMs so the temporal correlations in the conflicts are not explored, and the EVT method can only be applied to a site-level safety analysis instead of an individual real-time crash estimation.

The causal model is another non-crash-based method where the crash outcome $y$ of an event depends on its initial condition $u$ and action $x$ shown as Figure~\ref{fig:causal}. The probability distribution of crash outcome is given by Equation~\ref{eq:causal}: \useshortskip
\begin{align} \label{eq:causal}
    p(y, x, u)=(y| x, u) p(x | u) p(u)
\end{align}
where $p(u)$ is the probability distribution of the initial condition and $p(x|u)$ is the conditional probability distribution of action under the initial condition. The crash probability is by summing the probabilities of all the actions that could lead to a crash  \citep{davis2011outline}. The model can also lead to a natural interpretation of the counterfactual element in the definition of conflict and \citep{yamada2019new} combined the causal model and the potential outcome model \citep{pearl2009causal} to create a traffic conflict measure that can quantify the effectiveness of a given evasive action taken by a driver to avoid crashes. However, there are lots of assumptions for this causal model such as defining a set of initial conditions $U$ and a set of evasive actions $X$. It is complicated to estimate the probability distribution for all possible evasive actions and initial condition and the studies that employ this definition usually focus on a small subset of possible interactions and participants \citep{arun2021systematic}.

We can conclude that the core idea behind both non-crash-based methods, EVT and causal model, are probability density estimation. EVT tries to estimate the distribution of the extreme values of surrogate safety measures like TTC while causal model tries to estimate the distribution of initial condition $u$, the conditional distribution of action $x$ as well as the condtional distribution of crash outcome $y$. Both methods reqire many constraints and assumptions. 
\subsection{Probabilistic Time Series Prediction}
Deep learning has been proven to be powerful in many tasks like prediction and sample generation when there is sufficient training data and computing resources \citep{lecun2015deep}. In the field of deep unsupervised learning, many generative models like the Flow model \citep{ho2019flow}, variational auto-encoder (VAE) \citep{kingma2013auto} and GAN \citep{goodfellow2014generative} have been used for density estimation and sample generation. Their main applications are in image generation and text generation. Combined with neural network structures like LSTM and transformer that can deal with time series data, these generative models can estimate density functions for each time step. Many different types of model structures \citep{rasul2020multivariate, rasul2021autoregressive, tang2021probabilistic} have been tested on several multivariate time series datasets from the UCI benchmark \citep{blake1998uci}. Many of these generative models use the autoregressive network \citep{germain2015made} to improve their model performance since the network imposes a dependency relationship between the previous estimated variable and the current estimated variable.
\section{Methodology}
\begin{figure}[H]
	\centering
	\includegraphics[width=0.7\linewidth]{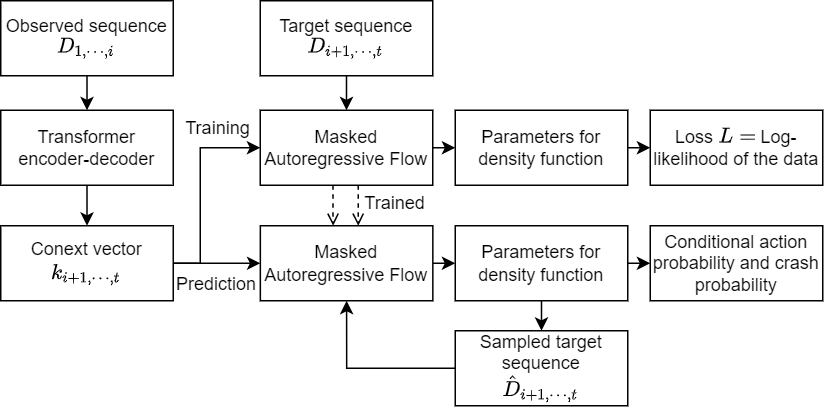}
	\caption{Workflow of the model}
	\label{fig:flowchart}
\end{figure}

In this section, we propose a framework to train the transformer-MAF model and sample sequence and calculate action and crash probability from the trained model. The diagram of the framework is summarized and visualized in Figure~\ref{fig:flowchart}. The data we use are high dimensional time series data containing $v_i$, the speed for vehicle $i$, $v_j$, the speed for vehicle $j$, $a_i$, the longitudinal acceleration for vehicle $i$,$a_j$ the longitudinal acceleration for vehicle $j$, and $TTC$. We define the condition $u = (v_i, v_j)$, the action $x = (a_i, a_j)$ and crash outcome $y = TTC$. Therefore, each data point $D$ contains a sequence of $ (u_{t},x_{t},y_{t})$ $\forall t= 1, \cdots, T$ where $t$ represents the current time step and $T$ is the sequence length. We did not use the distance between two vehicles in the condition $u$ because the previous $TTC$ and speed in the observed sequence could imply the distance information.

During training, the high dimensional time series data $D$ is divided into an observed sequence and a target sequence. The observed sequence is fed into the transformer encoder-decoder model and the model outputs the latent representation $k$ which will be called as the context vector in this paper. We then pass the context vector $k$ and the current time step data $(u_t, x_t, y_t)$ together into the MAF model. The parameters of the density function for time $t$ are estimated for each data and each time step and used to calculate the log-likelihood of the data which is the loss function.

During prediction, the trained transformer-MAF is used. Only the observed sequence is fed into the model and the context vector $k$ is computed as in the training phase. Using the context vector $k$, the MAF model estimates the parameters for the density function at time $t$ and samples the current time step data $(\hat{u}_t, \hat{x}_t, \hat{y}_t)$ autoregressively. Finally the density functions are used to computes the conditional action probability, the crash probability and conditional crash probability.
\subsection{Masked Autoregressive Flow Model}
In this section, we will first introduce the 1-dimensional flow model and then extend it into a high dimensional flow model where masked autoregressive flow is one of its variants. First, we will explain why flow model is better in our use case compared to other deep unsupervised model like VAE and GAN. The goal of this research is to do density estimation for the SSM data $a = (u, x, y)$ and calculate the action and crash probability. Therefore, we need to explicitly estimate a density function $p(u,x,y)$ for the SSM data while VAE and GAN model do not meet this requirement. Another advantage of the flow model is that it can deal with continuous variables easily. One common objective for density estimation is to maximize the log likelihood of the data as Equation~\ref{eq:max_log}, however, condition $u$, action $x$ and TTC $y$ are all continuous variables and it is difficult to ensure two conditions in Equation~\ref{eq:conds} for a proper distribution since a continuous variable can take any value in a range. \useshortskip

\begin{align}
&\max _{\theta} \sum_{i} \log p\left(a^{(i)}\right) \label{eq:max_log}\\
\int_{-\infty}^{+\infty} p(a) \,da & = 1 \quad \text{ and } \quad p(a) > 0 \quad \forall a \label{eq:conds}
\end{align}
However, flow model can transform these unknown and complex continuous random variables $(u,x,y)$ into other random variables $(z_1,z_2,z_3)$ with known and simple probability density functions $p_{Z_1}$, $p_{Z_2}$ and $p_{Z_3}$. A common choice for them is standard Gaussian distribution \citep{dinh2016density}. We will use $z_1 = f_{\theta}(u)$ as an example where $f_{\theta}$ is the transformation. With the change of variable theorem, we have Equation~\ref{eq:l1}-~\ref{eq:l3} and therefore the loss function becomes easy to compute given that $p_{Z_1}$ is standard Gaussian. There are two requirements for the transformation $f_{\theta}$. One is differentiability since we need to calculate the derivative of $f_{\theta}$ and the other one is invertibility since we need to calculate $\hat{u}$ from $f^{-1}_{\theta}$. Fortunately, using a deep neural network as the transformation $f_{\theta}$ satisfies these two requirements with suitable activation functions \citep{papamakarios2021normalizing}. During sampling, we can sample $z_1 \sim N(0,1)$ and then calculate $\hat{u} = f^{-1}_{\theta}(z_1)$. 

\begin{align}
\int p(u) d u &=\int p_{Z_1}(z_1) d z_1 = 1 \label{eq:l1}\\
p(u) =p_{Z_1}(z_1)|\frac{dz_1}{du}| &= p_{Z_1}\left(f_{\theta}(u)\right)\left|\frac{\partial f_{\theta}(u)}{\partial u}\right| \label{eq:l2}\\
\max _{\theta} \sum_{i} \log p\left(u^{(i)}\right) =\max _{\theta} \sum_{i} &\log p_{Z_1}\left(f_{\theta}\left(u^{(i)}\right)\right)+\log \left|\frac{\partial f_{\theta}}{\partial u}\left(u^{(i)}\right)\right| \label{eq:l3}
\end{align}

For high dimensional data, the training process is the same and researchers usually add an autoregressive structure between each dimension to increase the performance of the neural network \citep{papamakarios2017masked, huang2018neural}. According to the causal model from \citep{davis2011outline}, there is a dependency relationship among condition $u$, action $x$ and crash outcome $y$ (in our case TTC) shown in Figure~\ref{fig:causal}. Therefore, adding this autoregressive structure can not only improve the model performance but also impose some physical meanings on the model and increase its interpretability. With this structure, we can calculate the conditional action probability, crash probability and conditional crash probability which are shown in the next sub-section. The transformation between $(u,x,y)$ and $(z_1,z_2,z_3)$ are shown in Equation~\ref{eq:l4}-\ref{eq:l6}. We will first sample $z_1$ to get $\hat{u}$ and then sample $z_2$ and combine with $\hat{u}$ to get $\hat{x}$ and lastly sample $z_3$ and combine with $\hat{u}$ and $\hat{x}$ to get $\hat{y}$. The sampling process has to been done step by step and it becomes slow for much higher dimensional data.\useshortskip

\begin{align}
\text{Training: }    &z_1 = f_{\theta}(u) \qquad \qquad \qquad \qquad \qquad \text{Sampling: } \hat{u}=f_{\theta}^{-1}\left(z_{1}\right) \label{eq:l4}\\
&z_2 = f_{\theta}(x;u)  \qquad \qquad \qquad \qquad \qquad \qquad \qquad \! \hat{x}=f_{\theta}^{-1}\left(z_{2} ; \hat{u}\right) \label{eq:l5}\\
&z_3 = f_{\theta}(y;x,u) \qquad \qquad \qquad \qquad \qquad \qquad \quad  \hat{y}=f_{\theta}^{-1}\left(z_{3} ; \hat{x}, \hat{u}\right) \label{eq:l6}
\end{align}

\begin{figure}[H]
    \centering
    \begin{subfigure}{0.65\textwidth}
        \centering
        \includegraphics[width=\linewidth]{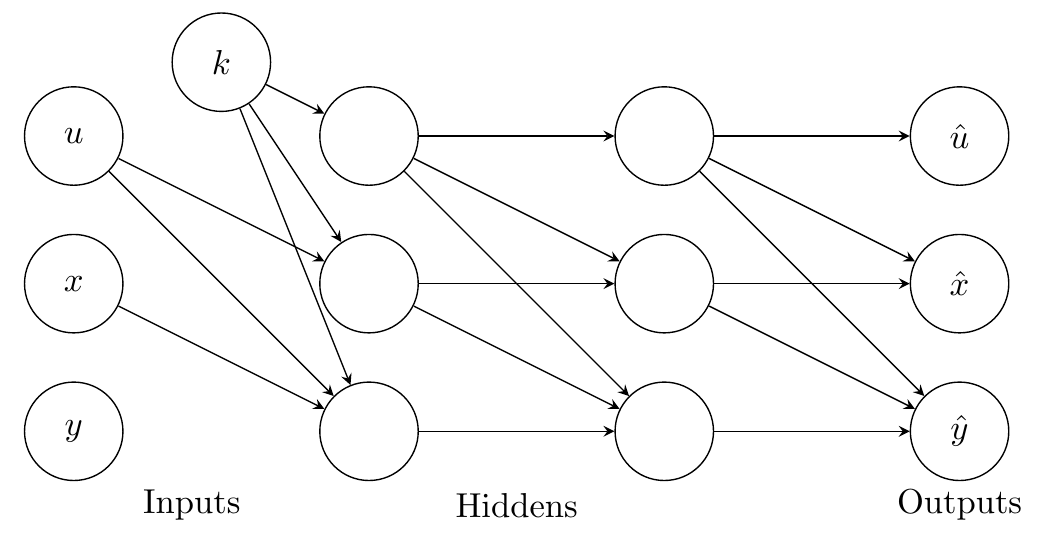}
	\caption{Connections between neurons under autoregressive structure}
	\label{fig:neuron_auto}
    \end{subfigure}%
    \begin{subfigure}{0.35\textwidth}
        \centering
        \includegraphics[width=0.59\linewidth]{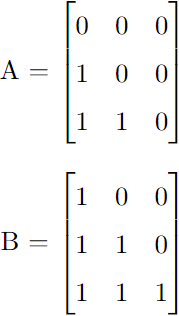}
	\caption{Masks applied on a fully connected network to ensure dependency structure}
	\label{fig:two_masks}
    \end{subfigure}%
\caption{Design autoregressive neural network for the SSM data}\label{fig:design}
\end{figure}

Based on \citep{germain2015made}, we design an autoregressive neural network shown as Figure~\ref{fig:design} to ensure the dependency structure among $(u,x,y)$. $k$ is the context vector from the Transformer encoder-decoder and all output neurons should have its information. Mask $A$ is applied on the fully connected network between inputs and the first hidden layers and mask $B$ is applied on the two hidden layers and the outputs. They guarantee the information from the inputs flows properly into the outputs.
\subsection{Action Prediction and Crash Probability Calculation}
The transformer encoder-decoder and the autoregressive flow model allow us to explicitly estimate three probability density function $p_t(u|k)$, $p_t(x|u,k)$ and $p_t(y|x,u,k)$ for each future time step $t$, we can calculate the conditional action probability $P_t(X|U,k)$, crash probability $P_t(y\leq 0|k)$ and conditional crash probability $P_t(y \leq 0 | X_t, U_t,k)$ where $X$ and $U$ represent some events of the random variables $x$ and $u$ and $k$ is the observed context vector. Since all the density functions contain time step $t$ and the context vector $k$, we will omit them in the following calculation for simplicity but keep in mind that different context vectors can drastically change the density functions which will be shown in the next section. 

The conditional action probability is the conditional probability of some action $X$ can happen under some condition $U$ which is shown in Equation~\ref{eq:l7}. We can evaluate this probability over any interval of action and under any interval of condition. For example, we can define evasive action as acceleration within range $[-6,-3] m/s^2$ and no action as acceleration within range $[-0.5,0.5] m/s^2$ and the 
probability of the driver doing evasive action and the probability of the driver doing no action under some condition $U$ can be calculated by replacing the corresponding intervals.

\begin{align} 
    P(x \in X | u \in U) = \frac{\int_{X}\int_{U}p(x|u)p(u)\,dx\,du}{\int_{U}p(u)\,du} \label{eq:l7}
\end{align}

The crash probability is the marginal probability of the crash outcome exceeding its critical point, in our case $TTC \leq 0$. This marginal probability can be written as the joint probability of a crash happens and all the possible actions $x$ and conditions $u$ are considered since $P(x \in \mathbb{R}) = 1$ and $P(u \in \mathbb{R}) = 1$. Introducing $x$ and $u$ can help us calculate the crash probability because our transformer-MAF model does not directly estimate the probability density of crash outcome $p(y)$ but the conditional probability density of crash outcome given action and condition $p(y|x,u)$. The calculation is shown in Equation~\ref{eq:no_ap}. However, in practice, the values of actions like acceleration can not span the entire real number set and we shrink the integration interval to the minimum and maximum values in the empirical data. And the condition $u$ like speed can only change slightly between each time step ($0.1s$ for our data) so we replace the real number set with the range from the 25th-percentile value to the 75-th percentile value $[U_{\text{25th}}, U_{\text{75th}}]$ as Equation~\ref{eq:ap}.

\begin{align} 
    P(y \leq 0) = P(y\leq 0, x\in \mathbb{R}, u \in \mathbb{R}) &= \int_{-\infty}^0\int_{R}\int_{R} p(y|x,u)p(x|u)p(u) \,dy\,dx\,du \label{eq:no_ap}\\
    &\approx \int_{-\infty}^0\int_{X_{\text{min}}}^{X_{\text{max}}}\int_{U_{\text{25th}}}^{U_{\text{75th}}} p(y|x,u)p(x|u)p(u) \,dy\,dx\,du \label{eq:ap}
\end{align}

The conditional crash probability is the conditional probability of a crash can happen given some action $X$ is taken and under some condition $U$ shown in Equation~\ref{eq:cond_c}. For a traffic conflict event, we can calculate the conditional crash probability if the no action was taken and compare it to the conditional crash probability given evasive action was taken by replacing the corresponding intervals. With this counterfactual experiment, we can explore the causal effect of the action on the crash outcome of an event.

\begin{align} 
    P(y \leq 0 | x \in X, u \in U) = \frac{\int_{-\infty}^0\int_{X}\int_{U} p(y|x,u)p(x|u)p(u) \,dy\,dx\,du}{\int_{X}\int_{U}p(x|u)p(u)\,dx\,du} \label{eq:cond_c}
\end{align}

Since we only know the probability density functions $p(u)$, $p(x|u)$ and $p(y|x,u)$, we can use the Monte Carlo method \citep{geweke1989bayesian} to calculate these integrals. However, the traditional Monte Carlo integration method can only work with finite intervals and can not handle the interval for $y$ $(-\infty, 0]$ which contains infinite limit. We use the change of variable method to map this interval to some finite interval and multiply the derivative to the integrand so that the traditional Monte Carlo method works properly \citep{cameron_mci}. The derivation of Equation~\ref{eq:l7}-\ref{eq:cond_c} can be found in Appendix~\ref{appendix:a}.

\section{Experiments and Results}
\subsection{Model training and evaluation}
The dataset we use for the study is INTERACTION dataset \citep{zhan2019interaction} which is extracted from videos of three unsignalized intersections and five roundabouts taken by drones and traffic cameras. The dataset records the position $(x,y)$, velocity $(v_x, v_y)$ and yaw angle $\theta$ for each vehicle within videos at every time step of $0.1$ second. Using the raw data, we calculate the longitudinal speed and acceleration and TTC with the constant speed assumption. We also fix the sequence length for each interaction to be 20 time steps (2 sec) where the first 10 (1 sec) is the observed sequence and the last 10 (1 sec) is the target sequence. The total shape of the entire dataset after processing is $(55055, 20, 5)$ which means there are $55055$ interactions as data points and each data point contains $5$ sequences with length $20$.

\begin{figure}[H]
    \centering
    \begin{subfigure}{0.5\textwidth}
        \centering
        \includegraphics[width=\linewidth]{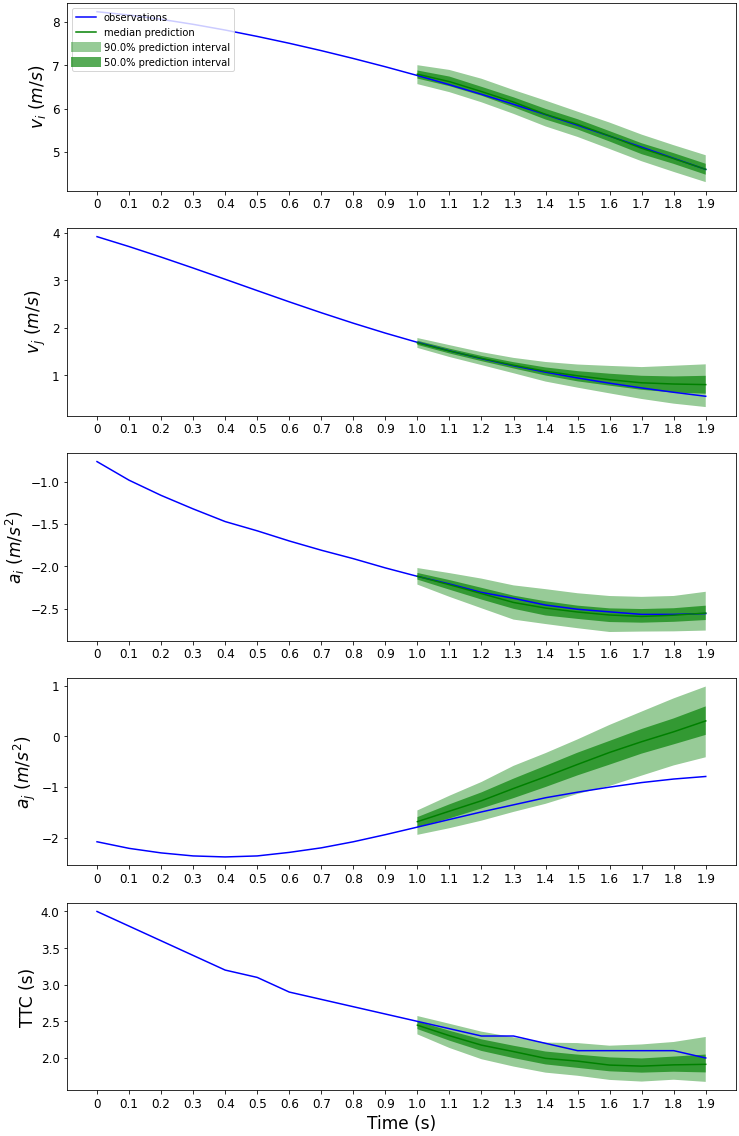}
        \caption{Interaction with low TTC and large deceleration\\(Traffic conflict)}
        \label{fig:22}
    \end{subfigure}%
    \begin{subfigure}{0.5\textwidth}
        \centering
        \includegraphics[width=\linewidth]{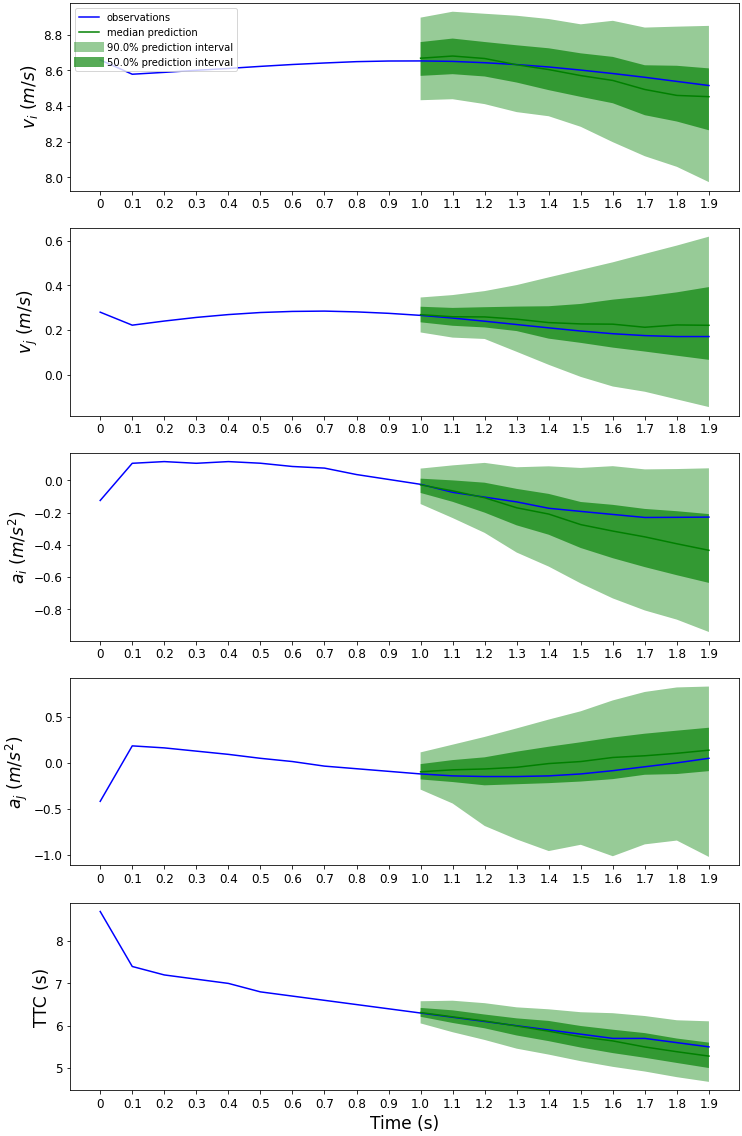}
        \caption{Interaction with high TTC and no deceleration\\(Normal Interaction)}
        \label{fig:221}
    \end{subfigure}%
\caption{Example of probabilistic SSM prediction}\label{fig:exam_SSM_pred}
\end{figure}

The entire dataset is split into training, validation and testing set by 80/10/10 and the model is trained on the training dataset with Adam optimizer and evaluated on validation set after every epoch. The hyperparameters for the model are shown in Table~\ref{tab:hyper} and the training and validation loss plot is shown in Figure~\ref{fig:loss_plot}. We select the model based on the best validation loss and use mean square error (MSE) and continuous ranked probability score (CRPS) to measure the accuracy of the trained model on the test set. For each test data, we sample 1000 sequences from the model. For MSE, we calculate the median value of the 1000 samples and compute the mean square error between the median and the ground truth sequence. For CRPS, it is defined as Equation~\ref{eq:ir} where $F$ is the cumulative function of random variable $X$ and $x$ is the observation \citep{matheson1976scoring} and it is often used to measure the difference between the predicted cumulative distribution function (CDF) and the empirical CDF of the observation. We use the weighted quantile loss to approximate CRPS similar to other implementations in Python package gluonts \citep{gluonts_jmlr} and properscoring \citep{gneiting2007strictly}. The final testing MSE is $0.17$ and CRPS is $0.1$.

\begin{align}\label{eq:ir}
    \text{CRPS}(F,x) = \int_{-\infty}^{\infty} (F(y) - \mathds{1}(y\geq x)^2)\,dy
\end{align}
\begin{table}[H]
\begin{minipage}[t]{0.5\linewidth}
\vspace{0pt}
\caption{Hyperparameters for \\transformer MAF model}
\centering
\begin{tabular}{cc}
\hline
Parameter   & Value     \\ \hline
Activation Function   & Gelu      \\ \hline
Model Dimension       & 40       \\ \hline
Feedforward Dimension & 160       \\ \hline
Dropout Rate          & 0.1       \\ \hline
Num. Attention Heads  & 8        \\ \hline
Num. Encoder Blocks   & 3         \\ \hline
Num. Decoder Blocks   & 3         \\ \hline
Positional Encoding   & Learnable \\ \hline
Num. Layers in MAF & 2           \\ \hline
Learning Rate         & 1e-3      \\ \hline
\end{tabular}
\label{tab:hyper}
\end{minipage}
\begin{minipage}[t]{0.5\linewidth}
\vspace{40pt}
\centering
	\includegraphics[width=\linewidth]{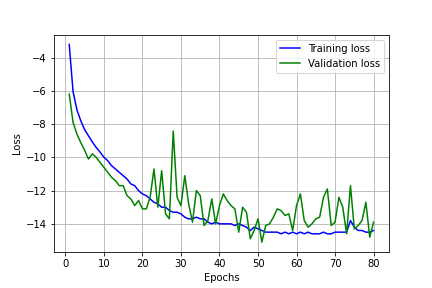}
	\captionof{figure}{Training \& Validation Loss}
	\label{fig:loss_plot}
\end{minipage}
\end{table}
Some visualizations of ground truth vs sampled data are shown in Figure~\ref{fig:exam_SSM_pred}. Since our data is multidimensional time series data, each visualization contains 5 sub-figures. X axis is time and Y axis represents a feature. The blue line is the ground truth data while the dark green line is the median of the prediction. The dark green area is 50\% prediction interval which is bound by the 25-percentile and 75-percentile predictions and the light green area is 90\% prediction interval which is bound by the 5-percentile and 95-percentile predictions. The bound area increases as time increases since the next prediction is based on the observed sequence and the pass predictions and the uncertainty accumulates through time. The visualizations show that the model can predict the future SSMs reasonably well and more visualizations can be found in the Appendix~\ref{appendix:b}.
We also implemented a non-autoregressive structure and compare the MSE and CRPS on the test set of two models. The non-autoregressive structure is shown in Figure~\ref{fig:neuron_nonauto}. Compared to the autoregressive structure in Figure~\ref{fig:neuron_auto}, this structure removes the connections from the input $u$ and $x$ to the hidden neurons so that there is no information in $u$ and $x$ passed to the outputs. The hidden layers and the output layers are fully connected since there is no constraint on the context information and it can also increase the model performance. In Table~\ref{tab:auto}, the result shows that the autoregressive structure can fit the data better which validates the dependency structure between $u$ and $x$ as well as $x$ and $y$.
\begin{table}[H]
\begin{minipage}[t]{0.5\linewidth}
\vspace{0pt}
\centering
	\includegraphics[width=\linewidth]{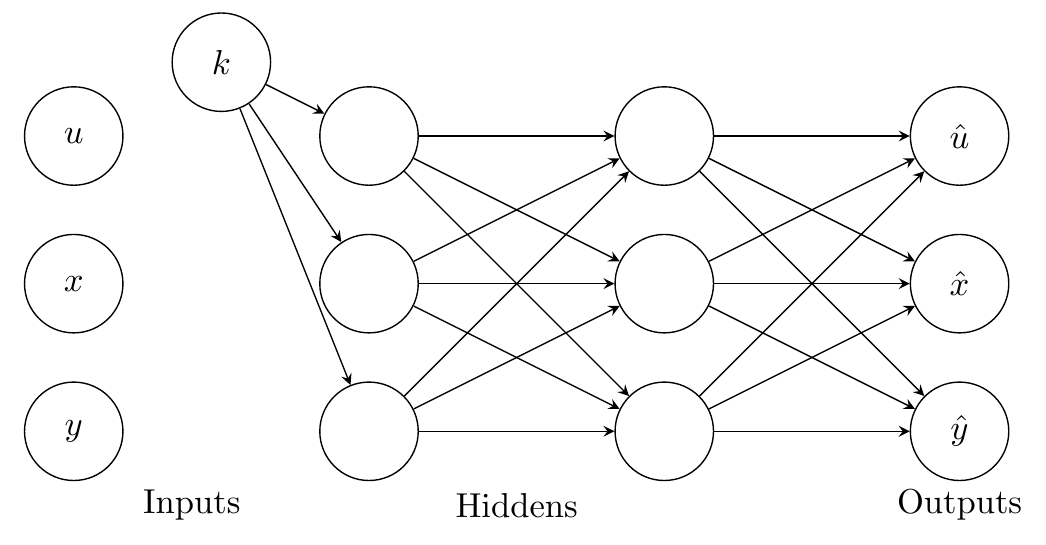}
	\captionof{figure}{Connections between neurons under non-autoregressive structure}
	\label{fig:neuron_nonauto}
\end{minipage}
\begin{minipage}[t]{0.5\linewidth}
\vspace{25pt}
\caption{MSE and CRPS of the autoregressive and\\ non-autoregressive structure (the lower the better)}
\centering
\begin{tabular}{ccc}
\hline
     & Autoregressive & \begin{tabular}[c]{@{}c@{}}Non-\\ autoregressive\end{tabular} \\ \hline
MSE  & 0.17           & 0.24                                                          \\ \hline
CRPS & 0.1            & 0.19                                                          \\ \hline
\end{tabular}
\label{tab:auto}
\end{minipage}
\end{table}

\subsection{Probability Prediction}
First we calculate the conditional action probability under the conditional $u \in [U_{\text{25th}}, U_{\text{75th}}]$ and two different contexts: traffic conflict and normal interaction. We define 5 actions with their names and value ranges shown in Table~\ref{tab:actions}. $X_{\text{eva}}$ represents an event when vehicle $i$ is doing evasive action and vehicle $j$ is doing all possible actions and similarly $X_{\text{no}}$ represents an event when vehicle $i$ is not taking any action and vehicle $j$ is doing all possible actions. We only consider vehicle $j$ doing all possible actions because there will be too many combinations with vehicle $j$ taking 5 different actions. More importantly, from vehicle $i$'s perspective, it does not know what action vehicle $j$ will take so it has to optimize its action with all possible actions from vehicle $j$ considered.
\begin{table}[H]
\centering
\caption{Notations and Definitions for Different Actions}
\begin{tabular}{cc}
\hline
Action            & Range \\ \hline
Evasive Action  $X_{\text{eva}}$   &   $(a_i \in [-6,-3], a_j \in [-6,6])$    \\ \hline
Large Deceleration &     $(a_i \in [-3,-2], a_j \in [-6,6])$  \\ \hline
Small Deceleration &    $(a_i \in [-2,-0.5], a_j \in [-6,6])$   \\ \hline
No Action   $X_{\text{no}}$       &     $(a_i \in [-0.5,0.5], a_j \in [-6,6])$  \\ \hline
Acceleration       &    $(a_i \in [0.5,6], a_j \in [-6,6])$   \\ \hline
\end{tabular}
\label{tab:actions}
\end{table}

\begin{figure}[H]
    \centering
    \begin{subfigure}{0.5\textwidth}
        \centering
        \includegraphics[width=\linewidth]{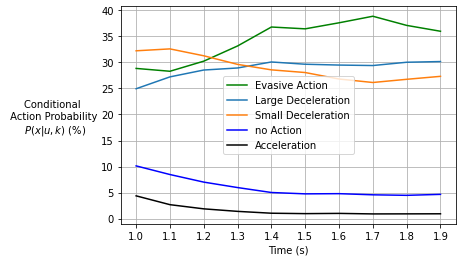}
        \caption{Traffic conflict context}
        \label{fig:cap_22}
    \end{subfigure}%
    \begin{subfigure}{0.4\textwidth}
        \centering
        \includegraphics[width=\linewidth]{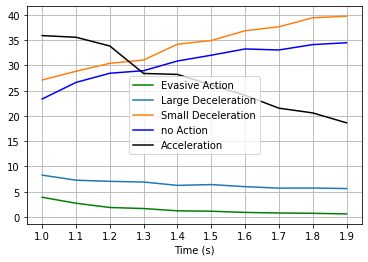}
        \caption{Normal Interaction context}
        \label{fig:cap_221}
    \end{subfigure}
\caption{Conditional action probability vs. time under different contexts}\label{fig:cap_d}
\end{figure}

\begin{figure}[H]
    \centering
    \begin{subfigure}{0.5\textwidth}
        \centering
        \includegraphics[width=0.95\linewidth]{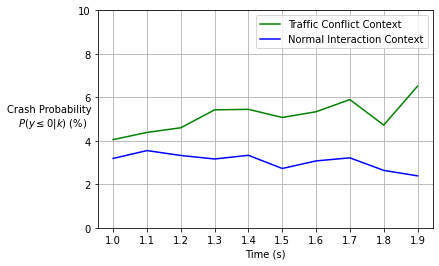}
        \caption{Crash probability vs. time}
        \label{fig:cp_100}
    \end{subfigure}%
    \begin{subfigure}{0.5\textwidth}
        \centering
        \includegraphics[width=\linewidth]{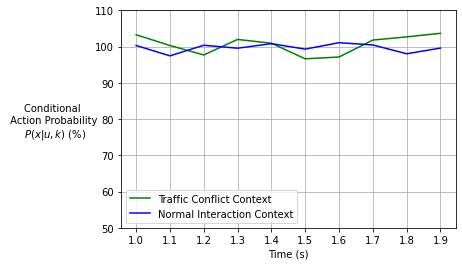}
        \caption{Conditional action probability vs. time}
        \label{fig:cap_100}
    \end{subfigure}%
\caption{Considering all possible actions under different scenarios}\label{fig:100}
\end{figure}

\begin{figure}[H]
    \centering
    \begin{subfigure}{0.5\textwidth}
        \centering
         \includegraphics[width=\linewidth]{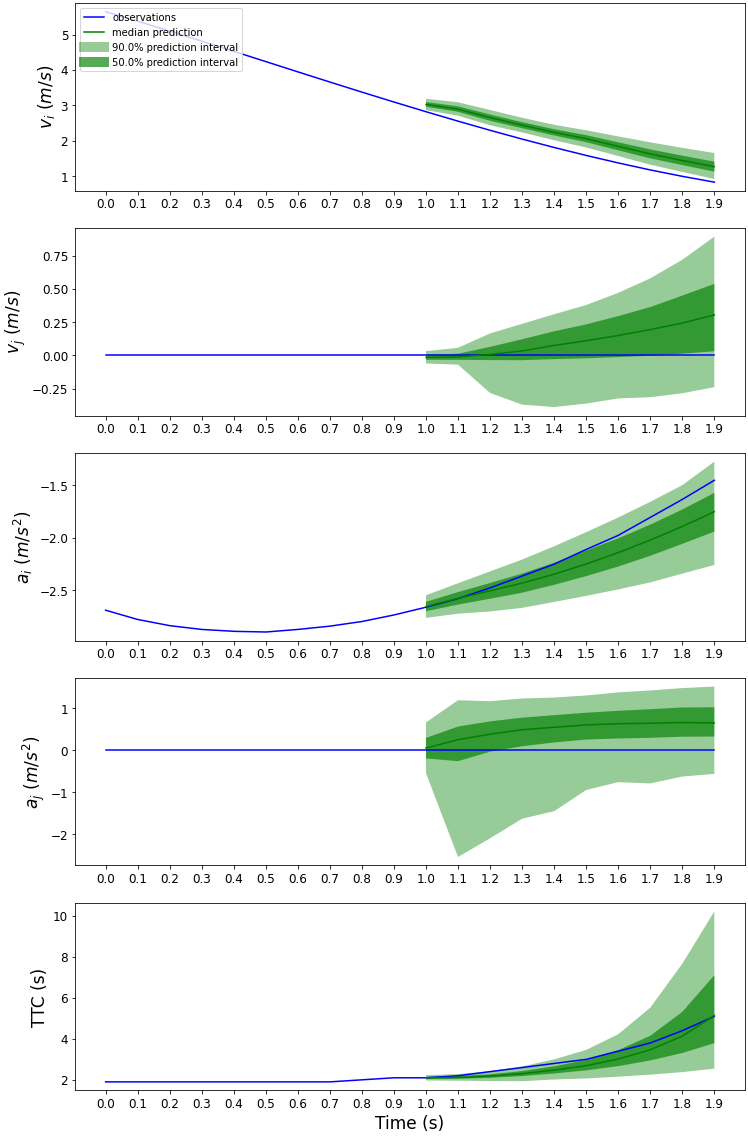}
        \caption{Correct prediction: evasive action}
        \label{fig:cf_cp}
    \end{subfigure}%
    \begin{subfigure}{0.5\textwidth}
        \centering
        \includegraphics[width=\linewidth]{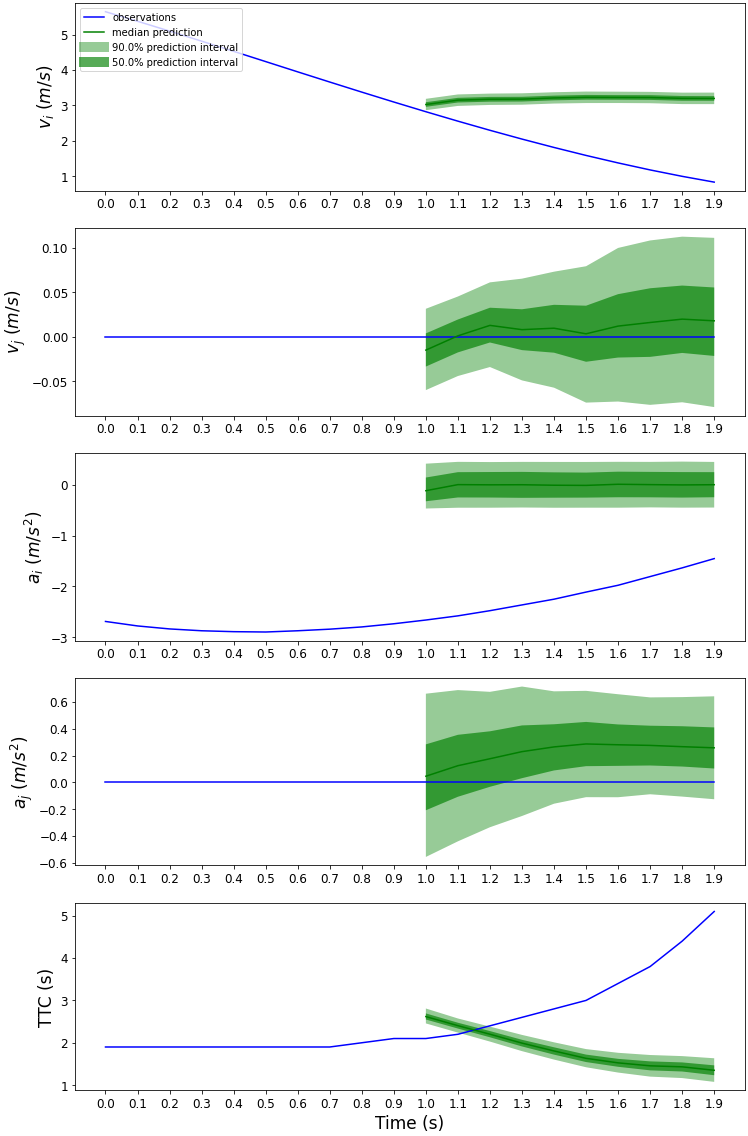}
        \caption{Counterfactual prediction: no action}
        \label{fig:cf_cfp}
    \end{subfigure}%
\caption{Correct prediction vs. counterfactual prediction on a traffic conflict context}\label{fig:cf_p}
\end{figure}

\begin{figure}[H]
    \centering
    \begin{subfigure}{0.5\textwidth}
        \centering
        \includegraphics[width=\linewidth]{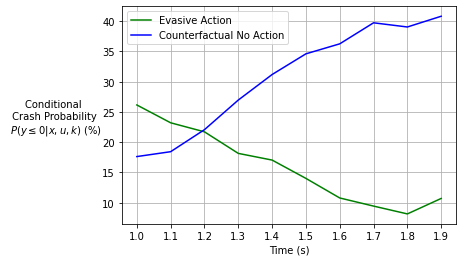}
        \caption{Conditional crash probability vs. time}
        \label{fig:cf_c}
        
    \end{subfigure}%
    \begin{subfigure}{0.5\textwidth}
        \centering
        \includegraphics[width=\linewidth]{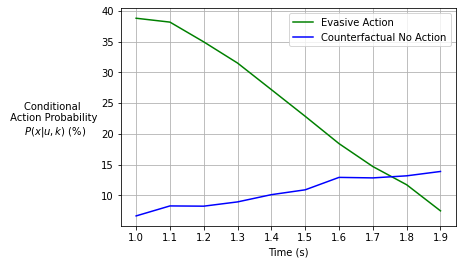}
        \caption{Conditional action probability vs. time}
        \label{fig:cf_a}
    \end{subfigure}%
\caption{Counterfactual example}\label{fig:cf}
\end{figure}

The probability of 5 different actions are evaluated in Figure~\ref{fig:cap_d}. We can validate these probabilities by summing them at each time step and the sum should be close to $100\%$ since these 5 actions cover the entire range of action values. Under the traffic conflict context in Figure~\ref{fig:22}, the model predicts that there will be high probability for evasive action and large deceleration (over $30\%$ and $25\%$) and low probability for no action and acceleration (below $10\%$ and $5\%$) to happen in the future 10 time steps. The probability for evasive action gradually grows as time increases. This aligns with the ground truth and the sampled sequence of $a_i$ in Figure~\ref{fig:22} since the actual and predicted speed values are decreasing over time. On the other hand, under the context of normal interaction in Figure~\ref{fig:221}, the model predicts that there will be high probability for no action and small deceleration (over $25\%$ and $30\%$) and low probability for evasive action and large deceleration (below $5\%$ and $10\%$) to happen in the future 10 time steps. The probability of acceleration decreases drastically (from $40\%$ to $20\%$) and the mass is shifted toward small acceleration and no action whose probabilities both increase around $10\%$. This aligns with the sampled sequence of $a_i$ in Figure~\ref{fig:221} because $90\%$ of the predicted values are within $[-0.2,0.1]$ at time $1.0$ and $75\%$ of the predicted values are within $[-0.9, -0.2]$ at time $1.9$.
We also calculate the crash probability under different contexts using Equation~\ref{eq:ap}. In Figure~\ref{fig:cp_100}, the model predicts the crash probability under traffic conflict context increases from $4\%$ to $6\%$ over time since the predicted TTC values in Figure~\ref{fig:22} are decreasing and getting closer and closer to $0$. On the other hand, the crash probability for normal interaction context is around $2.5\%$ over the entire future $10$ steps since the predicted TTC values in Figure~\ref{fig:221} are high and far away from $0$. We also calculate the conditional action probability with $x \in X_{\text{all}} = (a_i \in [-6,6], a_j \in [-6,6])$ in Figure~\ref{fig:cap_100} and they are around $100\%$ for both contexts which also validates the correctness of the Monte Carlo integration process since theoretically $P(x \in \mathbb{R} | U,k) = 1$.
We select a rear-end conflict in Figure~\ref{fig:cf_cp} where vehicle $i$ is approaching to vehicle $j$ from behind and vehicle $j$ is already stopped as our counterfactual experiment. In the counterfactual scenario, vehicle $i$ didn't perform evasive action but instead take no action from time $1.0$ to time $1.9$ (flat green zone in $a_i$ of Figure~\ref{fig:cf_cfp}), therefore the model is forced to take this "fake" no action as input to predict the future condition and future TTC as Figure~\ref{fig:cf_cfp}. The predicted $v_i$ becomes a flat green zone with small variation since the counterfactual $a_i$ is small and can only take values in $[-0.5, 0.5]$. The model also predicts that the future TTC will continue to decrease if vehicle $i$ takes no action. The conditional crash probabilities $P(y\leq 0 | x \in X, u \in [U_{\text{25th}}, U_{\text{75th}}])$ for evasive action $X_{\text{eva}}$ and no action $X_{\text{no}}$ are shown in Figure~\ref{fig:cf_c}. The result shows that evasive action is effective to avoid crashes because when the driver is doing evasive action, the conditional crash probability decreases over time while if the driver was doing no action, the conditional crash probability would increase over time. Moreover, in this counterfactual experiment, only the action taken by the driver is changed but not any other variables like condition $u$ or context $k$.

The conditional crash probability is a lot higher than the crash probability and this is because in Equation~\ref{eq:cond_c}, the conditional crash probability is the ratio of the crash probability and the action probability which are both small. Shown as the green curve in Figure~\ref{fig:cf_a}, the model predicts the needs for evasive action drops over time which matches the $a_i$ curve in Figure~\ref{fig:cf_cp}. On the other hand, the blue curve in Figure~\ref{fig:cf_a} shows the model initially predicts that counterfactual no action should not be taken (around $5\%$ probability) but slowly increases this probability to $15\%$ because a series of the counterfactual no actions are taken by the driver and those actions become a new context which affects the future prediction.

\subsection{Discussion}
Our trained transformer-MAF model can sample accurate future sequences and estimate the conditional action probability, crash probability and conditional crash probability reasonably. Under the traffic conflict context, the model estimates a high probability ($40\%$) of evasive action while under the normal interaction context, the model estimates a high probability ($35\%$) of no action. This conditional action probability is achieved by masking the last $5$ time steps of the observed action sequence ($x_{6,\cdots,10}$). Without this masking mechanism, the conditional action probability can reach $90\%$ for $X_{\text{eva}}$ under traffic conflict context which is problematic because predicting $x_{11}$ given $x_{1,\cdots,10}$ is trivial and the context vector $k$ would be dominated by the information of $x_{1,\cdots,10}$ \citep{zerveas2020transformer}. The current action should not completely depend on its previous actions since in the context of traffic conflict, evasive action is often taken abruptly \citep{johnsson2018search} and it should also depend on the perceived safety by the driver which means the driver knows that he is driving fast (condition $u$) and perceives that he might crash soon (TTC $y$). Therefore, we mask out the $x_{6, \cdots, 10}$ to reduce the information from action sequence so that the model can learn a context with more information extracted from the observed condition and TTC sequence. Similar to the observed action sequence, we also mask the last $5$ time steps of the observed TTC sequence. Without the mask, the context vector $k$ contains too much information of the previous TTC and the model will create a density function $p(y|x,u,k)$ where $x$ and $u$ are ignored and it becomes $p(y|k)$. This means that no matter what current action $x_t$ and current $u_t$ are, the density function of $p(y|x,u,k)$ will not change and it fails our purpose to build the causal model among $u,x$ and $y$ as well as estimate the counterfactual probability. However, if we mask out the entire observed action sequence or TTC sequence, the predicted density function and the sampled sequence will be off compared to the ground truth. Therefore we learn that there is a trade-off between the accuracy and variety of the predicted density function and we need to find the balance point so that the prediction has enough variation and high accuracy.

The crash probability under traffic conflict context is around $5\%$ and under normal interaction context is around $3\%$. These values are a little bit higher than those reported in \citep{zheng2019bayesian, borsos2020collision} (around $3\%$). The reason is that we sub-sampled the original processed data. The entire processed data contains all the interactions that have a possible collision course and a lot of them $(50\%)$ have relatively high minimum TTC (greater than $4s$) over the possible collision course. Therefore, we removed those interactions with minimum TTC greater than $4s$ from the data and created a smaller dataset. This is good from a model training standpoint because the portion of interactions with low TTC $(<2.5s)$ is changed from $10\%$ to $20\%$ after the sub-sampling and it becomes easier for the model to capture the pattern of these minority interactions. On the other hand, the subsampling process also changed the underlying distribution of the entire dataset and it can result in a biased estimation of the density function. One way to ease this inflated crash probability problem could be multiplying the crash probability by $50\%$ to counter the sub-sampling effect. We can also create a larger model that can directly work with the entire dataset or use other methods like re-weighting the loss to handle imbalanced class problem.

The results of the time series conditional action prediction as well as crash prediction can be applied in many cases. In a real time driving scenario, vehicles with advanced driver assistance sytem (ADAS) \citep{kukkala2018advanced} or connected and autonomous vehicles (CAV) \citep{elliott2019recent} can capture sequences of conditions $u$ like speed and location and actions $x$ like acceleration and steering of themselves and other surrounding objects and then calculate the crash outcome $y$ like TTC. These data can be directly fed into our transformer-MAF model to predict a sequence of future conditional action probability and crash probability. The predicted action and crash probability can be incorporated into the collision avoidance system. In a scenario of safety assessment, we can select the max value of the predicted crash probability sequence as the crash probability of an interaction. For each site, we can calculate the average crash probability of the interactions from this site as the crash probability of the site and prioritize sites based on this value. Moreover, the average conditional evasive action probability can indicate whether large deceleration is often used in the interactions from a site which can give some insight for the safety investigators. It is important to note that the actions and the corresponding ranges defined in Table~\ref{tab:actions} can be changed under different driving scenarios as well as the crash interval $[-\infty,0]$. For example, the crash interval can be changed to $[-\infty,1]$ if the system needs a safer buffer when calculating the probability. This can be done without retraining the model because because the model directly estimates a continuous probability density function of action $x$ which provides flexibility on probability inference. If the model was trained on a discrete variable with pre-defined thresholds for different actions, the entire model needs to be retrained if the thresholds change.

\begin{figure}[H]
	\centering
	\includegraphics[width=0.5\linewidth]{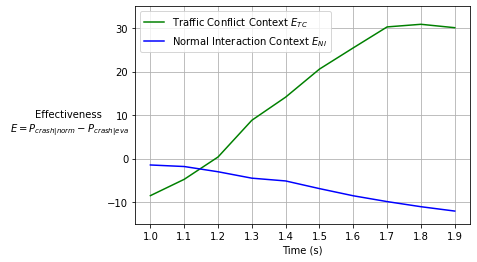}
	\caption{Effectiveness of evasive action under different context}
	\label{fig:effect}
\end{figure}

The conditional crash probability allows us to test the counterfactual situation since it is defined as the crash probability given action $X$ is taken and under condition $U$. With different counterfactual action given as input, we can calculate different conditional crash probability. Specifically, we can calculate the conditional crash probability $P_{\text{crash}|\text{no}} = P(y \leq 0 | X_{\text{no}}, U, k)$ if the driver had taken no action and the conditional crash probability $P_{\text{crash}|\text{eva}} = P(y \leq 0 | X_{\text{eva}}, U, k)$ if the driver had taken an evasive action under traffic conflict context. According to the potential outcome model, We define $E = P_{\text{crash}|\text{no}} - P_{\text{crash}|\text{eva}}$ to quantify the effectiveness of the evasive action to avoid crashing. The effectiveness under traffic conflict context in Figure~\ref{fig:cf_p} defined as $E_{\text{TC}}$ is a lot higher than the effectiveness under normal interaction context in Figure~\ref{fig:cf_n} defined as $E_\text{NI}$ which is shown in Figure~\ref{fig:effect}. This is logical because we would expect a high effect of evasive action in a traffic conflict context and no effect in a normal interaction context. The counterfactual prediction plot of the normal interaction and all probability tables can be found in Appendix~\ref{appendix:c}.
\section{Conclusion}
This paper proposed a method to connect surrogate safety measures to crash probability via causal probabilistic time series prediction. The density functions of condition, action and crash outcome are estimated for each time step using transformer-MAF. The causal relationship among those variables are implemented in the neural network with autoregressive structure. The conditional action probability, crash probability and conditional crash probability are calculated based on the estimated density functions. The results show that the sampled SSM sequences are accurate by measuring the MSE and CRPS. The estimated probabilities are comparable to those reported in the literature. Moreover, the effectiveness of evasive action to avoid crashes is evaluated with the potential outcome model.

Our method overcomes the limitations of the causal model with the help of deep learning and traffic data collection techniques. There are lots of improvements can be done to the current method. More variables such as relative distance between two vehicles can be added to condition $u$, steering can be added to action $x$ and other types of crash outcome like PET can be incorporated as well. The current method can only predict the density function up to future 10 time steps and a larger model can be explored to estimate longer or variable-length sequences.
\appendix
\begin{appendices}
\section{Equation Derivation}\label{appendix:a}
Conditional action probability in Equation~\ref{eq:l7}:
\begin{align*} 
    P(x \in X | u \in U) & = \frac{P(x \in X , u \in U)}{P(u \in U)}\\
    & = \frac{\int_{X}\int_{U}p(x,u)\,dx\,du}{\int_{U}p(u)\,du}\\
    &= \frac{\int_{X}\int_{U}p(x|u)p(u)\,dx\,du}{\int_{U}p(u)\,du}
\end{align*}

Crash probability in Equation~\ref{eq:no_ap}:
\begin{align*}
    P(y \leq 0, x \in R, u \in R) & = \int_{-\infty}^0\int_{R}\int_{R} p(y,x,u) \,dy\,dx\,du \\
    & = \int_{-\infty}^0\int_{R}\int_{R} p(y|x,u) p(x,u)\,dy\,dx\,du\\
    &= \int_{-\infty}^0\int_{R}\int_{R} p(y|x,u)p(x|u)p(u) \,dy\,dx\,du
\end{align*}

Conditional crash probability in Equation~\ref{eq:cond_c}:
\begin{align*} 
    P(y \leq 0 | x \in X, u \in U) &=\frac{P(y \leq 0, x \in X, u \in U)}{P(x \in X, u \in U)}\\
    &= \frac{\int_{-\infty}^0\int_{X}\int_{U} p(y|x,u)p(x|u)p(u) \,dy\,dx\,du}{\int_{X}\int_{U}p(x|u)p(u)\,dx\,du}
\end{align*}
\section{More SSM Probabilistic Prediction Plots} \label{appendix:b}
\begin{figure}[H]
\centering
    \begin{subfigure}{0.5\textwidth}
        \centering
        \includegraphics[width=\linewidth]{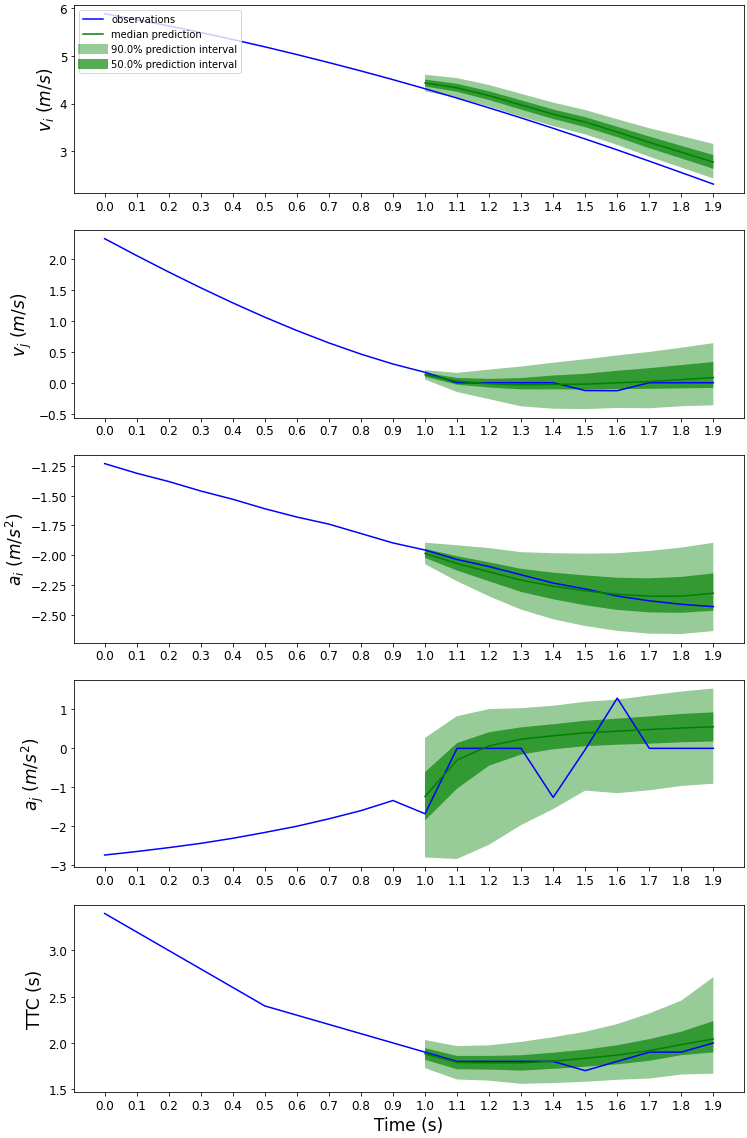}
        \caption{}
    \end{subfigure}%
    \begin{subfigure}{0.5\textwidth}
        \centering
        \includegraphics[width=\linewidth]{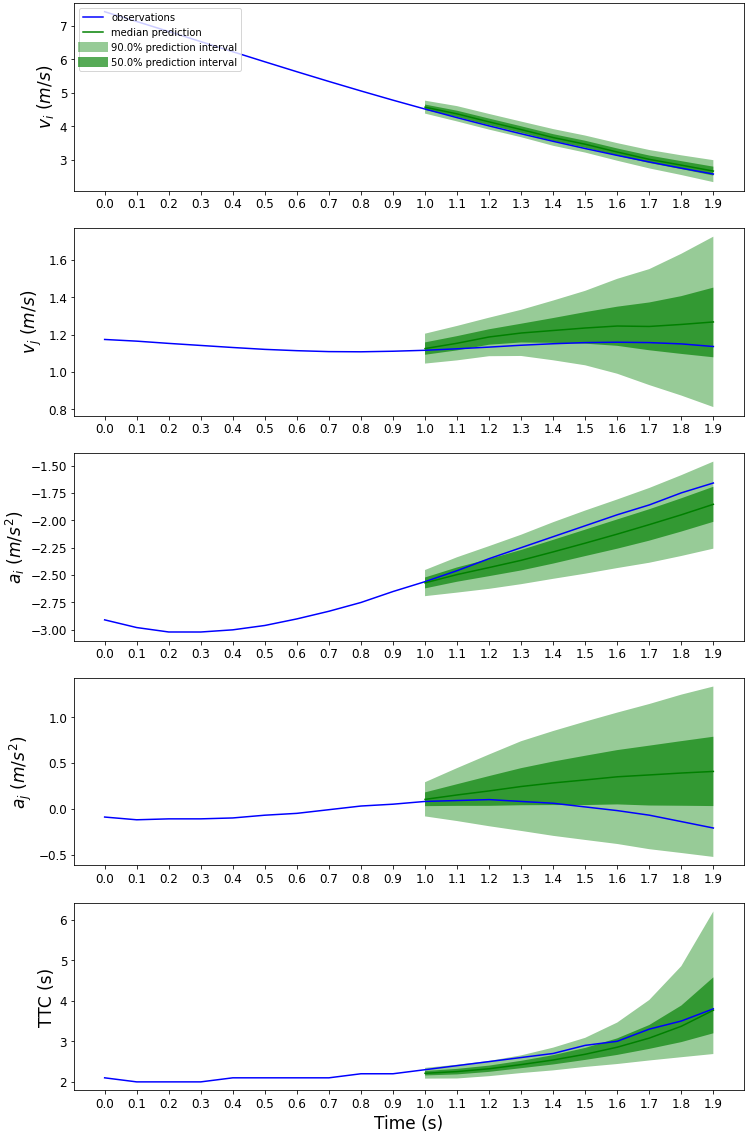}
       \caption{}
    \end{subfigure}
\end{figure}
\begin{figure}[H]
\ContinuedFloat 
\begin{subfigure}[b]{0.5\textwidth}
        \centering
        \includegraphics[width=\linewidth]{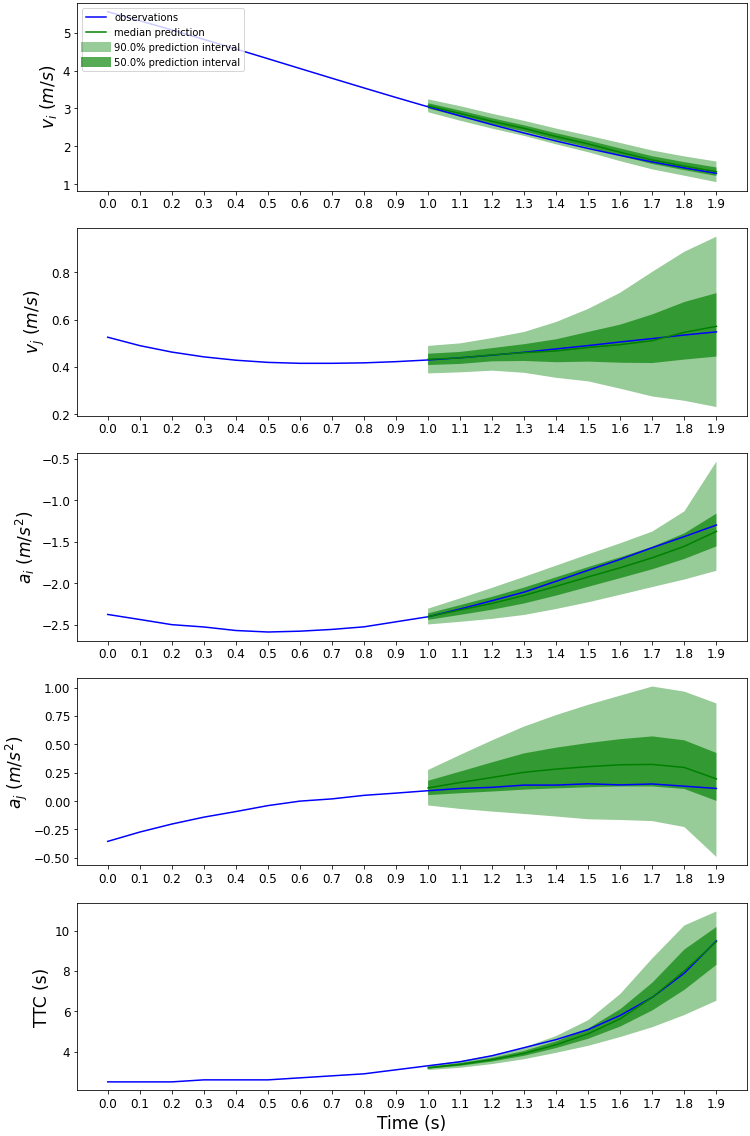}
        \caption{}
    \end{subfigure}%
    \begin{subfigure}[b]{0.5\textwidth}
        \centering
        \includegraphics[width=\linewidth]{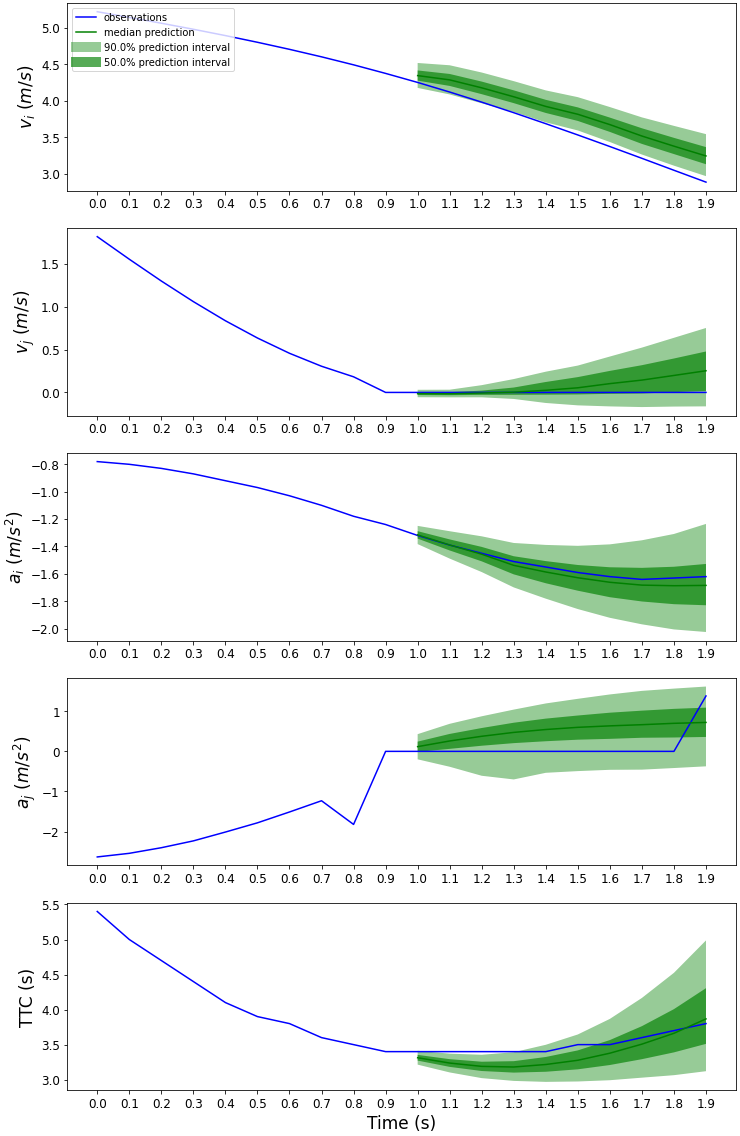}
        \caption{}
    \end{subfigure}
    \end{figure}
\begin{figure}[H]
    \ContinuedFloat 
    \begin{subfigure}[b]{0.5\textwidth}
        \centering
        \includegraphics[width=\linewidth]{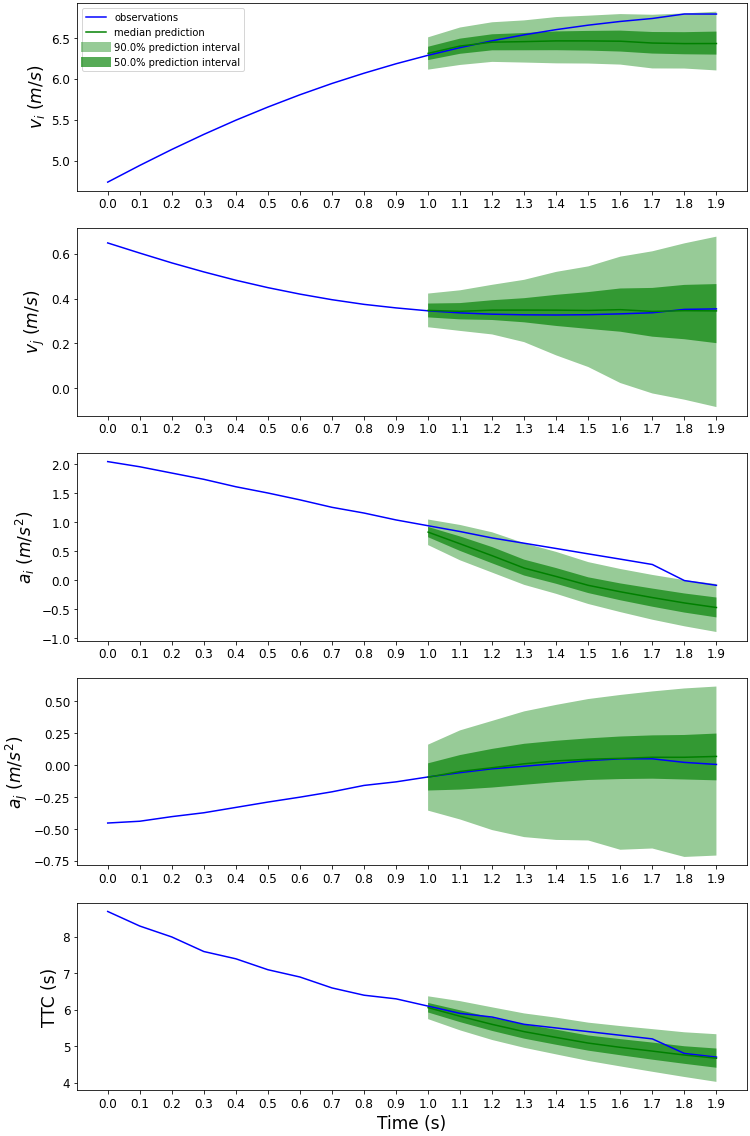}
        \caption{}
    \end{subfigure}%
    \begin{subfigure}[b]{0.5\textwidth}
        \centering
        \includegraphics[width=\linewidth]{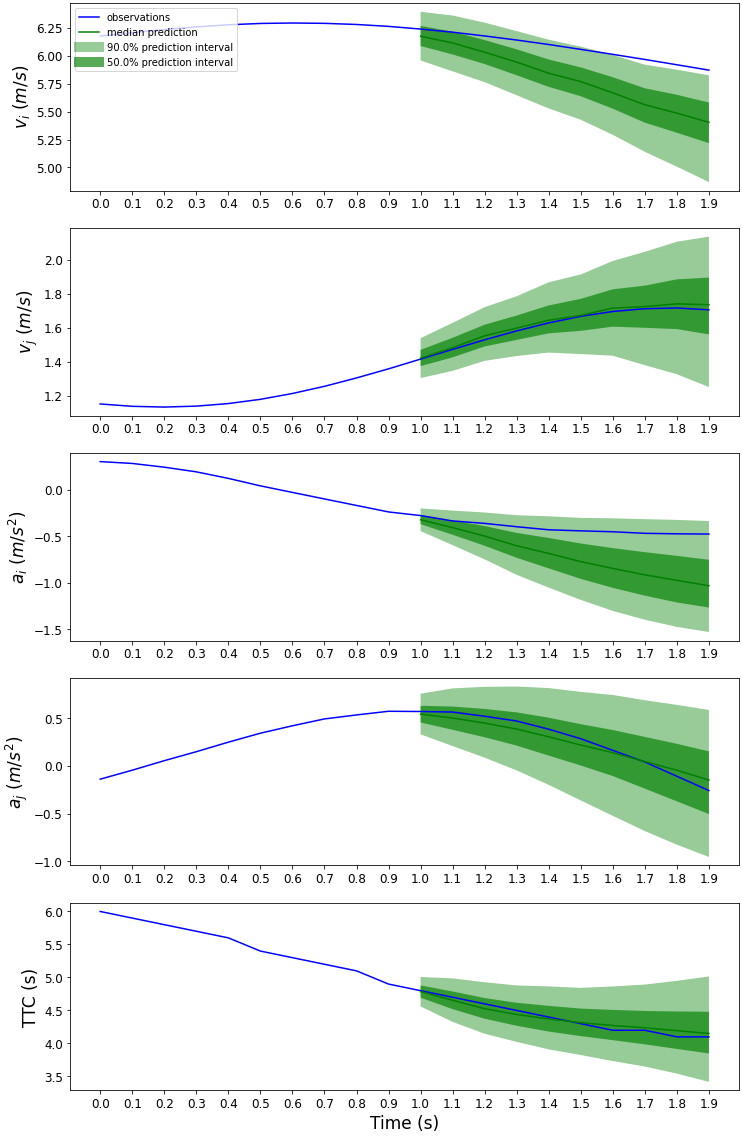}
        \caption{}
    \end{subfigure}
\end{figure}
\begin{figure}[H]
    \ContinuedFloat 
    \begin{subfigure}[b]{0.5\textwidth}
        \centering
        \includegraphics[width=\linewidth]{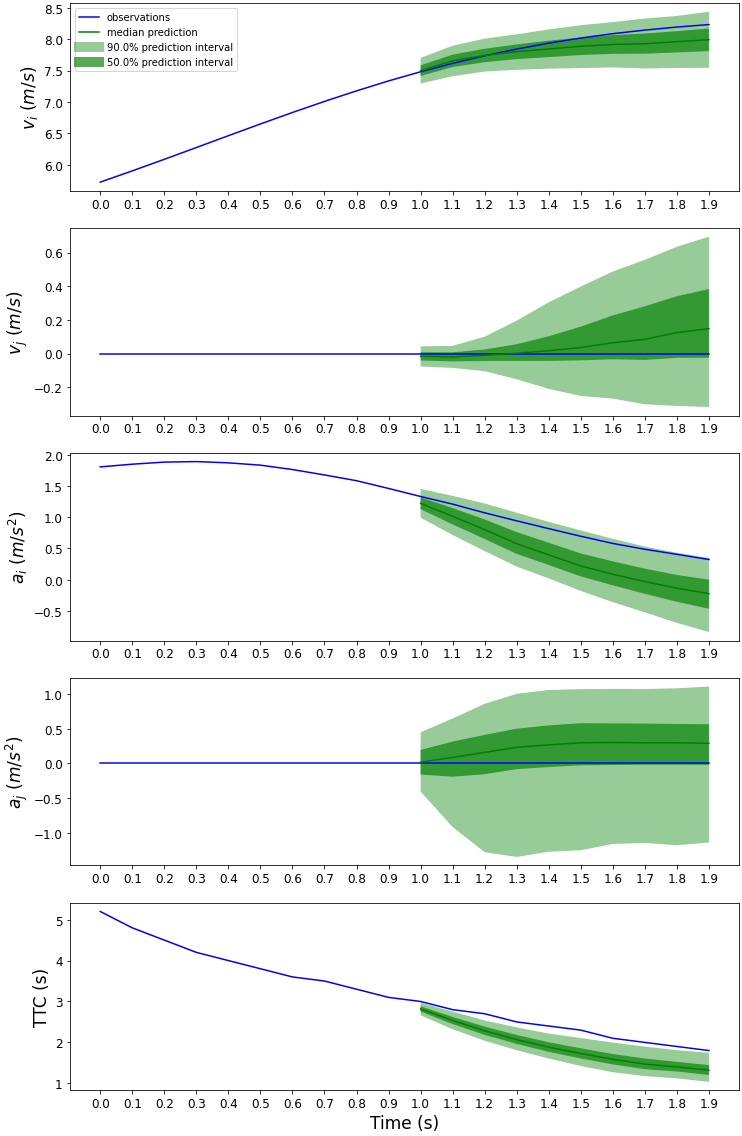}
        \caption{}
    \end{subfigure}%
    \begin{subfigure}[b]{0.5\textwidth}
        \centering
        \includegraphics[width=\linewidth]{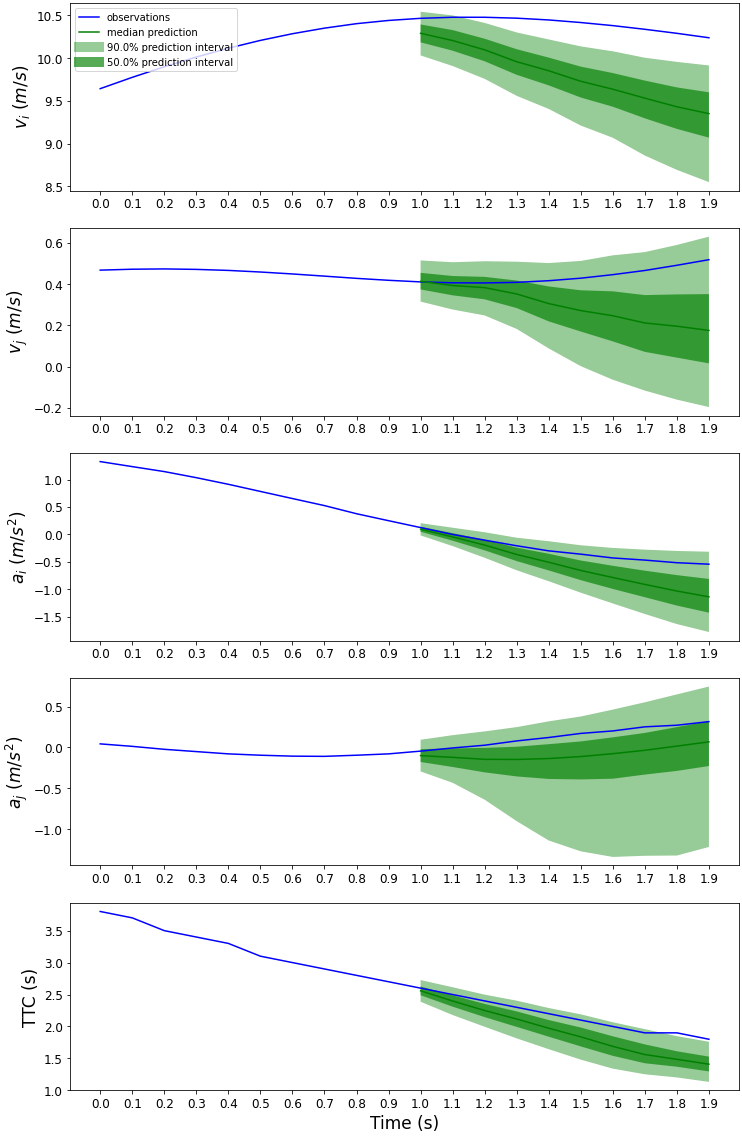}
       \caption{}
    \end{subfigure}
    \caption{More Examples of probabilistic SSM prediction}
\end{figure}

\section{Raw Data for the Probability}\label{appendix:c}
Table~\ref{tab:22_other_ac} and~\ref{tab:221_other_ac} are used to generate Figure~\ref{fig:cap_d}.
\begin{table}[H]
\centering
\caption{Conditional Action Probability ($\%$) under \\context $k=$ traffic conflict at Figure~\ref{fig:22} and $U = \{u \in U_{50\%}\}$}
\begin{tabular}{ccccccccccc}
\hline
                                                             & 1.0 & 1.1 & 1.2 & 1.3 & 1.4 & 1.5 & 1.6 & 1.7 & 1.8 & 1.9 \\ \hline
\begin{tabular}[c]{@{}c@{}}Evasive\\ Action\end{tabular}  &28.85 & 28.31 & 30.25 & 33.17 & 36.81 & 36.46 & 37.62 & 38.88 & 37.12 & 35.99 \\ \hline
\begin{tabular}[c]{@{}c@{}}Large\\ Deceleration\end{tabular} &24.95 & 27.23 & 28.54 & 28.94 & 30.09 & 29.66 & 29.50 & 29.41 & 30.04 & 30.17 \\ \hline
\begin{tabular}[c]{@{}c@{}}Small\\ Deceleration\end{tabular} &32.24 & 32.61 & 31.28 & 29.63 & 28.58 & 28.06 & 26.82 & 26.15 & 26.76 & 27.33 \\ \hline
\begin{tabular}[c]{@{}c@{}}No\\ Action\end{tabular}  & 10.12 & 8.46 & 7.00 & 5.96 & 5.01 & 4.75 & 4.79 & 4.56 & 4.46 & 4.65 \\ \hline
Acceleration                     &  4.36 & 2.67 & 1.87 & 1.38 & 1.03 & 0.95 & 0.99 & 0.89 & 0.90 & 0.92 \\ \hline
\end{tabular}\label{tab:22_other_ac}
\end{table}
\begin{table}[H]
\centering
\caption{Conditional Action Probability ($\%$) under \\context $k=$ normal interaction at Figure~\ref{fig:221} and $U = \{u \in U_{50\%}\}$}
\begin{tabular}{ccccccccccc}
\hline
                                                             & 1.0 & 1.1 & 1.2 & 1.3 & 1.4 & 1.5 & 1.6 & 1.7 & 1.8 & 1.9 \\ \hline
\begin{tabular}[c]{@{}c@{}}Evasive\\ Action\end{tabular} & 3.90 & 2.71 & 1.87 & 1.66 & 1.21 & 1.13 & 0.90 & 0.78 & 0.72 & 0.60 \\ \hline
\begin{tabular}[c]{@{}c@{}}Large\\ Deceleration\end{tabular} & 8.28 & 7.27 & 7.05 & 6.91 & 6.25 & 6.41 & 5.99 & 5.71 & 5.72 & 5.62 \\ \hline
\begin{tabular}[c]{@{}c@{}}Small\\ Deceleration\end{tabular} & 27.12 & 28.84 & 30.42 & 31.08 & 34.20 & 34.93 & 36.87 & 37.65 & 39.44 & 39.74 \\ \hline
\begin{tabular}[c]{@{}c@{}}No\\ Action\end{tabular}          &  23.36 & 26.64 & 28.45 & 28.96 & 30.88 & 32.02 & 33.26 & 33.06 & 34.12 & 34.49 \\ \hline
Acceleration                                                 &  35.90 & 35.58 & 33.85 & 28.41 & 28.22 & 26.12 & 24.09 & 21.54 & 20.60 & 18.64 \\ \hline
\end{tabular}\label{tab:221_other_ac}
\end{table}

The first two rows of Table~\ref{tab:22_all} and ~\ref{tab:221_all} are used to generate Figure~\ref{fig:100}.
\begin{table}[H]
\centering
\caption{Probability ($\%$) for context $k=$ traffic conflict at Figure~\ref{fig:22}\\ and $Y = \{y \leq 0\}, X = X_{\text{all}} = \{a_1 \in [-6,6], a_2 \in [-6,6]\}, U = \{U_{25\%}, U_{75\%}\}$}
\begin{tabular}{ccccccccccc}
\hline
         & 1 & 1.1 & 1.2 & 1.3 & 1.4 & 1.5 & 1.6 & 1.7 & 1.8 & 1.9 \\ \hline
$P(X|U)$   &103.3 & 100.3 & 97.70 & 102 & 100.9 & 96.63 & 97.13 & 101.8 & 102.7 & 103.7  \\ \hline
$P(Y,X,U)$ &4.06 & 4.39 & 4.60 & 5.43 & 5.45 & 5.07 & 5.34 & 5.90 & 4.72 & 6.51   \\ \hline
$P(X,U)$   &22.78 & 23.38 & 24.51 & 27.22 & 28.08 & 24.52 & 24.99 & 25.45 & 22.01 & 28.49    \\ \hline
$P(Y|X,U)$ &17.80 & 18.76 & 18.77 & 19.93 & 19.40 & 20.70 & 21.36 & 23.17 & 21.46 & 22.86   \\ \hline
\end{tabular}\label{tab:22_all}
\end{table}
\begin{table}[H]
\centering
\caption{Probability ($\%$) for context $k=$ normal interaction at Figure~\ref{fig:221}\\ and $Y = \{y \leq 0\}, X = X_{\text{all}} = \{a_1 \in [-6,6], a_2 \in [-6,6]\}, U = \{u \in [U_{25\%}, U_{75\%}\}]$}
\begin{tabular}{ccccccccccc}
\hline
         & 1 & 1.1 & 1.2 & 1.3 & 1.4 & 1.5 & 1.6 & 1.7 & 1.8 & 1.9 \\ \hline
$P(X|U)$   &100.3 & 97.45 & 100.4 & 99.56 & 100.8 & 99.30 & 101.1 & 100.5 & 98.02 & 99.59 \\ \hline
$P(Y,X,U)$ &3.19 & 3.55 & 3.33 & 3.16 & 3.33 & 2.73 & 3.08 & 3.22 & 2.64 & 2.39  \\ \hline
$P(X,U)$   &24.42 & 25.83 & 27.03 & 24.45 & 26.47 & 22.62 & 26.30 & 28.43 & 24.61 & 22.99    \\ \hline
$P(Y|X,U)$ &13.06 & 13.75 & 12.30 & 12.94 & 12.59 & 12.06 & 11.70 & 11.32 & 10.73 & 10.39   \\ \hline
\end{tabular}\label{tab:221_all}
\end{table}

The first and last rows of Table~\ref{tab:9_eva} and ~\ref{tab:9_norm_ct} are used to generate Figure~\ref{fig:cf}.
\begin{table}[H]
\centering
\caption{Probability ($\%$) for context $k=$ traffic conflict at Figure~\ref{fig:cf_cp}\\ and $Y = \{y \leq 0\}, X = X_{\text{eva}} = \{a_1 \in [-6,-3], a_2 \in [-6,6]\}, U = \{u \in U_{50\%}\}$}
\begin{tabular}{ccccccccccc}
\hline
         & 1 & 1.1 & 1.2 & 1.3 & 1.4 & 1.5 & 1.6 & 1.7 & 1.8 & 1.9 \\ \hline
$P(X|U)$   &38.78 & 38.15 & 34.94 & 31.49 & 27.17 & 22.84 & 18.40 & 14.67 & 11.71 & 7.47\\ \hline
$P(Y,X,U)$ &2.42 & 1.97 & 1.69 & 1.62 & 1.08 & 0.74 & 0.51 & 0.35 & 0.26 & 0.18  \\ \hline
$P(X,U)$   &9.25 & 8.47 & 7.78 & 8.92 & 6.32 & 5.28 & 4.72 & 3.66 & 3.13 & 1.65   \\ \hline
$P(Y|X,U)$ &26.15 & 23.21 & 21.71 & 18.15 & 17.05 & 14.02 & 10.79 & 9.45 & 8.16 & 10.71   \\ \hline
\end{tabular}\label{tab:9_eva}
\end{table}

\begin{figure}[H]
    \centering
    \begin{subfigure}{0.5\textwidth}
        \centering
         \includegraphics[width=\linewidth]{nons_plot_221.png}
        \caption{Correct prediction: driver is taking no action}
        \label{fig:cf_n1}
    \end{subfigure}%
    \begin{subfigure}{0.5\textwidth}
        \centering
        \includegraphics[width=\linewidth]{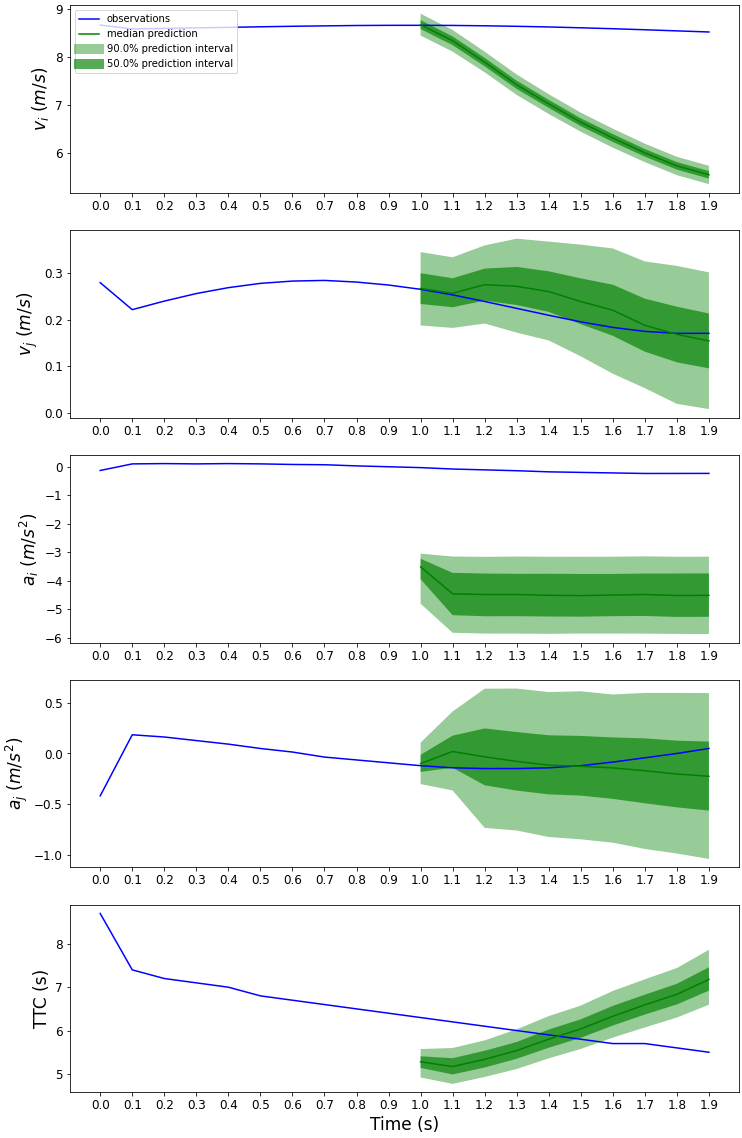}
        \caption{Counterfactual prediction: if the driver had taken evasive action}
        \label{fig:cf_n2}
    \end{subfigure}%
\caption{Correct prediction vs. counterfactual prediction under normal interaction context}\label{fig:cf_n}
\end{figure}

\begin{table}[H]
\centering
\caption{Counterfactual probability ($\%$) for context $k=$ traffic conflict at Figure~\ref{fig:cf_cp}\\ and $Y = \{y \leq 0\}, X = X_{\text{no}}= \{a_1 \in [-0.5,0.5], a_2 \in [-6,6]\}, U = \{u \in U_{50\%}\}$}
\begin{tabular}{ccccccccccc}
\hline
         & 1 & 1.1 & 1.2 & 1.3 & 1.4 & 1.5 & 1.6 & 1.7 & 1.8 & 1.9 \\ \hline
$P(X|U)$   &6.64 & 8.27 & 8.23 & 8.92 & 10.10 & 10.89 & 12.91 & 12.83 & 13.16 & 13.87\\ \hline
$P(Y,X,U)$ &0.34 & 0.34 & 0.48 & 0.67 & 0.86 & 0.85 & 1.08 & 1.28 & 0.97 & 1.43 \\ \hline
$P(X,U)$   &1.91 & 1.84 & 2.18 & 2.49 & 2.75 & 2.46 & 2.98 & 3.22 & 2.50 & 3.50 \\ \hline
$P(Y|X,U)$ &17.63 & 18.43 & 22.07 & 26.96 & 31.18 & 34.60 & 36.25 & 39.72 & 39.03 & 40.81 \\ \hline
\end{tabular}\label{tab:9_norm_ct}
\end{table}
Moreover, $E_{\text{TC}}$ is the results of the last row of Table~\ref{tab:9_norm_ct} minus the last row of Table~\ref{tab:9_eva} and $E_{\text{no}}$ is the results of the last row of Table~\ref{tab:221_norm} minus the last row of Table~\ref{tab:221_eva_ct}. And Table~\ref{tab:eff} generates Figure~\ref{fig:effect}.
\begin{table}[H]
\centering
\caption{Probability ($\%$) for context $k=$ normal interaction at Figure~\ref{fig:221}\\ and $Y = \{y \leq 0\}, X = X_{\text{no}} \{a_1 \in [-0.5,-0.5], a_2 \in [-6,6]\},  U = \{u \in U_{50\%}\}$}
\begin{tabular}{ccccccccccc}
\hline
         & 1 & 1.1 & 1.2 & 1.3 & 1.4 & 1.5 & 1.6 & 1.7 & 1.8 & 1.9 \\ \hline
$P(X|U)$   &23.36 & 26.64 & 28.45 & 28.96 & 30.88 & 32.02 & 33.26 & 33.06 & 34.12 & 34.49 \\ \hline
$P(Y,X,U)$ &0.66 & 0.78 & 0.80 & 1.01 & 1.15 & 0.99 & 0.93 & 0.87 & 0.87 & 1.03   \\ \hline
$P(X,U)$   &5.31 & 6.24 & 6.77 & 7.96 & 8.92 & 8.43 & 7.79 & 7.67 & 8.24 & 9.95    \\ \hline
$P(Y|X,U)$ &12.35 & 12.50 & 11.74 & 12.66 & 12.89 & 11.80 & 11.90 & 11.30 & 10.54 & 10.32   \\ \hline
\end{tabular}\label{tab:221_norm}
\end{table}
\begin{table}[H]
\centering
\caption{Counterfactual probability ($\%$) for context $k=$ normal interaction at Figure~\ref{fig:221}\\ and $Y = \{y \leq 0\}, X = X_{\text{eva}} = \{a_1 \in [-6,-3], a_2 \in [-6,6]\}, U = \{u \in U_{50\%}\}$}
\begin{tabular}{ccccccccccc}
\hline
         & 1 & 1.1 & 1.2 & 1.3 & 1.4 & 1.5 & 1.6 & 1.7 & 1.8 & 1.9 \\ \hline
$P(X|U)$   &3.76 & 23.98 & 12.36 & 11.90 & 11.94 & 11.62 & 11.48 & 10.38 & 10.00 & 9.63  \\ \hline
$P(Y,X,U)$ &0.12 & 0.85 & 0.48 & 0.57 & 0.51 & 0.57 & 0.53 & 0.50 & 0.61 & 0.54   \\ \hline
$P(X,U)$   &0.90 & 5.91 & 3.28 & 3.34 & 2.83 & 3.03 & 2.60 & 2.35 & 2.82 & 2.39    \\ \hline
$P(Y|X,U)$ &13.82 & 14.33 & 14.77 & 17.17 & 18.04 & 18.71 & 20.46 & 21.19 & 21.61 & 22.39   \\ \hline
\end{tabular}\label{tab:221_eva_ct}
\end{table}
\begin{table}[H]
\centering
\caption{Effectiveness of evasive action under traffic conflict and normal interaction contexts}
\begin{tabular}{ccccccccccc}
\hline
         & 1 & 1.1 & 1.2 & 1.3 & 1.4 & 1.5 & 1.6 & 1.7 & 1.8 & 1.9 \\ \hline
$E_{\text{TC}}$   &3.76 & 23.98 & 12.36 & 11.90 & 11.94 & 11.62 & 11.48 & 10.38 & 10.00 & 9.63  \\ \hline
$E_{\text{no}}$ &0.12 & 0.85 & 0.48 & 0.57 & 0.51 & 0.57 & 0.53 & 0.50 & 0.61 & 0.54   \\ \hline
\end{tabular}\label{tab:eff}
\end{table}
Additionally, the counterfactual plot for normal interaction is shown in Figure~\ref{fig:cf_n2}.
\end{appendices}
\bibliography{ref}

\end{document}